\documentclass[10pt, journal, final]{IEEEtran}  % Comment this line out if you need a4paper

\makeatletter

\let\proof\@undefined
\let\endproof\@undefined
\makeatother

\IEEEoverridecommandlockouts                              % This command is only needed if 

\usepackage{cite}
\usepackage[nocomma]{optidef}
\usepackage{amsmath,amssymb,amsfonts}
\usepackage{algorithm,algorithmicx}
\usepackage[noend]{algpseudocode}
\usepackage{calc}
\usepackage{subcaption}
\usepackage[font=footnotesize]{caption}
\usepackage{url}
\usepackage{booktabs}
\usepackage{multirow} 

\usepackage{enumitem,kantlipsum}
\usepackage{float}
\usepackage{graphics}
\usepackage{graphicx}
\usepackage{hyperref}
\hypersetup{
    colorlinks=true, 
    urlcolor=cyan,
    }
\usepackage{textcomp}
\usepackage{xcolor}
\usepackage{amsthm}  % for nice formatting of theorem env
\usepackage{bm}
\usepackage[nocomma]{optidef}

\theoremstyle{definition}
\newtheorem{theorem}{Theorem}
\newtheorem{problem}{Problem}

\newtheorem{observation}{Observation}
\newtheorem{definition}{Definition}
\newtheorem{remark}{Remark}

\algrenewcommand\algorithmiccomment[1]{\hfill// \text{\small #1}} % or use '//' instead of '\(\triangleright\)'

\DeclareMathOperator{\sign}{sgn}

\newcommand{\obs}{\mathcal{W}_{\text{obs}}}
\newcommand{\G}{\mathcal{G}}

\title{
AUTO-IceNav: A Local Navigation Strategy for Autonomous Surface Ships in Broken Ice Fields
}

\author{Rodrigue de Schaetzen$^{1}$, Alexander Botros$^{2}$, Ninghan Zhong$^{3}$, Kevin Murrant$^{4}$, Robert Gash$^{4}$, Stephen L.\ Smith$^{5}$% 
\thanks{This work is supported in part by the National Research Council Canada.} %
\thanks{$^{1}$Department of Computer Science and Operations Research, Université de
Montréal, Montréal, QC, Canada (e-mail: \protect\url{rodrigue.deschaetzen@mila.quebec})}
\thanks{$^{2}$Integrus Solutions (e-mail: \protect\url{alex.botros@integrus-solutions.com}).}
\thanks{$^{3}$Institute for Robotics and Intelligent Machines, Georgia Institute of Technology, Atlanta, GA 30332, USA (e-mail: \protect\url{nzhong34@gatech.edu}).}
\thanks{$^{4}$National Research Council Canada, St. John's, Newfoundland and Labrador, Canada (e-mail: \protect\url{{robert.gash, kevin.murrant}@nrc-cnrc.gc.ca}).}
\thanks{$^{5}$Department of Electrical and Computer Engineering, University of Waterloo, Waterloo, ON N2L 3G1, Canada (e-mail: \protect\url{stephen.smith@uwaterloo.ca}).}%
\thanks{Work done while RD, AB, and NZ were at the University of Waterloo.}
\thanks{Code and videos are available at {\color{magenta}\url{https://github.com/rdesc/AUTO-IceNav}}.}}

\begin{document}
\IEEEoverridecommandlockouts
\IEEEpubid{\begin{minipage}{\textwidth}\centering
\tiny
\copyright~2025 IEEE. Personal use of this material is permitted.  Permission from IEEE must be obtained for all other uses, in any current or future media, including 
reprinting/republishing this material\\for advertising or promotional purposes, creating new collective works, 
for resale or redistribution to servers or lists, or reuse of any copyrighted component of this work in other works.
\end{minipage}}

\maketitle
\pagestyle{empty}
\pagenumbering{gobble}

%%%%%%%%%%%%%%%%%%%%%%%%%%%%%%%%%%%%%%%%%%%%%%%%%%%%%%%%%%%%%%%%%%%%%%%%%%%%%%%%
\begin{abstract}
Ice conditions often require ships to reduce speed and deviate from their main course to avoid damage to the ship. In addition, broken ice fields are becoming the dominant ice conditions encountered in the Arctic, where the effects of collisions with ice are highly dependent on where contact occurs and on the particular features of the ice floes. In this paper, we present AUTO-IceNav, a framework for the autonomous navigation of ships operating in ice floe fields. Trajectories are computed in a receding-horizon manner, where we frequently replan given updated ice field data. During a planning step, we assume a nominal speed that is safe with respect to the current ice conditions, and compute a reference path. We formulate a novel cost function that minimizes the kinetic energy loss of the ship from ship-ice collisions and incorporate this cost as part of our lattice-based path planner. The solution computed by the lattice planning stage is then used as an initial guess in our proposed optimization-based improvement step, producing a locally optimal path. Extensive experiments were conducted both in simulation and in a physical testbed to validate our approach. 
\end{abstract}

\begin{IEEEkeywords}
Autonomous surface ships, ice navigation, marine vehicles, planning
\end{IEEEkeywords}

%%%%%%%%%%%%%%%%%%%%%%%%%%%%%%%%%%%%%%%%%%%%%%%%%%%%%%%%%%%%%%%%%%%%%%%%%%%%%%%%
\section{Introduction}

Ice navigation is becoming of significant interest to the maritime sector due to a number of factors. Commercial shipping, for instance, has increasingly used the Arctic shipping routes in the last decade due to a reduction in ice cover \cite{chircop2016sustainable}. The benefits of these alternative shipping routes include a decrease in emissions and a reduction in costs attributed to shorter travel times, reduced travel distances, and lower fuel consumption \cite{hansen2016arctic, li2020voyage, ari2013optimal}. There is also interest in improving supply to Northern communities, ferry operations, Arctic patrol, and environmental monitoring by sea \cite{quillerou2020arctic}. Despite these opportunities, the Arctic still presents considerable challenges and risks to safe and efficient navigation due to seasonally and regionally varying sea ice \cite{ryan2021arctic, canada_2019, kim2019numerical}. Sea ice conditions are affected by a range of different factors, both environmental (e.g., ocean currents and temperature) and operational (e.g., ship traffic and icebreaker assistance). This results in a wide variety of possible  sea ice characteristics that may be encountered during ship transit, each type presenting its own set of associated hazards~\cite{sabolic2022autonomous}.

\begin{figure}[!t]
    \centering
    \includegraphics[width=\columnwidth]{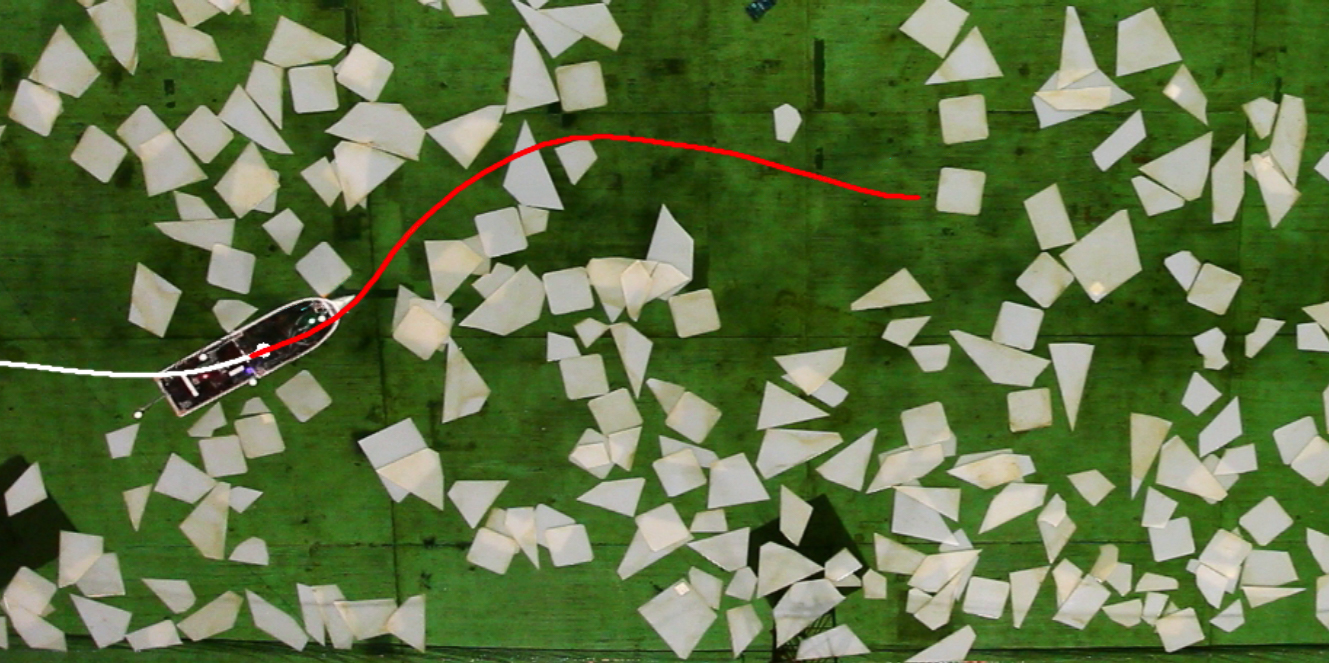}
    \caption{Snapshot of a sample trial from our physical experiments conducted in the Offshore Engineering Basin research facility in St. John’s, NL, Canada.}
    \label{fig:intro_fig}
\end{figure}
\IEEEpubidadjcol

Addressing the challenges of ice navigation fits well with the broader ongoing push toward greater ship autonomy and more intelligent systems to improve maritime safety and efficiency \cite{fossen2011handbook, veksler2016dynamic, murrant2021dynamic}. Although prior works have shown success in computing global routes that minimize ice navigation costs such as the average ice-induced forces exerted on the ship \cite{smith2022autonomous, lehtola2019finding}, none have considered the problem of optimizing navigation at the local level where costs are computed based on individual ship-ice collisions. In this work, we address the problem of local navigation for the autonomous maneuvering of surface ships in ice-covered waters, under the following key assumptions: (i) the environment consists of broken sea ice, (ii) the sea ice concentration---the fraction of a given area covered by sea ice---may be too high to achieve collision-free maneuvers, and (iii) the ship is rated for ice navigation but is not considered an icebreaker.

In contrast to level ice where continuous sheets are formed on the ocean surface requiring breaking of the ice, broken ice fields result in a much wider variety of possible ship-ice interactions \cite{ryan2021arctic}. Moreover, the most prevalent sea ice conditions expected to be encountered in the Arctic are ice floe fields \cite{li2022review, thomson2018overview}, and are the particular ice environments considered in this work (see Fig.~\ref{fig:intro_fig}). An ice floe field is defined as a collection of discrete floating ice pieces, called \emph{ice floes}, which may be characterized by their thickness, size, and shape. Differences among floes in a given ice field often vary greatly in terms of size and shape \cite{canada_2020, zhang2022semantic}. As a result, the collision responses vary significantly \cite{li2022review, lubbad2018simulator, daley2014gpu} and generally involve the ice floe being partially crushed and then pushed by the ship, leading to a loss of ship kinetic energy \cite{kim2019numerical}. The problem of local navigation in broken ice fields thus requires a method for efficiently finding a sequence of maneuvers that can be executed by the ship and are optimized in terms of the modeled costs of the ship-ice collisions. We provide a solution to this task by formulating a local planning problem in which collisions along a candidate planned path are evaluated from a kinetic energy loss perspective, based on the currently detected ice conditions.

In this paper, we extend our preliminary work~\cite{deschaetzen2023} in the following ways. First, we present an extension to our cost function using an ice concentration representation of the ice conditions to account for scenarios where an additional cost is incurred from pushing ice floes not in direct contact with the ship. Next, we propose an optimization-based improvement step as a second stage to our planning approach that produces a locally optimal solution to the continuous path planning problem. In contrast to simulation experiments from \cite{deschaetzen2023}, our experimental setup uses the dynamics of a full-scale vessel model, as well as a realistic distribution of ice floes and physics parameters. Our set of evaluation metrics includes statistics on the impact forces recorded during ship-ice collisions and the total energy consumption of the ship. The improved evaluation therefore captures a better picture of navigation performance, illustrating the trade-offs associated with changing course to avoid collisions with ice. In the physical experiments, we present quantitative results to complement the qualitative findings discussed in \cite{deschaetzen2023}. 
Finally, we release all code and documentation including our open source physics simulator for ship navigation in broken ice fields. 

\subsection{Contributions}

The following are the contributions of this work:
\begin{enumerate}
    \item A method for autonomous real-time ship navigation of broken ice fields called AUTO-IceNav (\textbf{Au}tonomous \textbf{T}wo-stage \textbf{O}ptimized \textbf{Ice} \textbf{Nav}igation) which produces desirable ship maneuvers that avoid head-on collisions with larger ice floes and areas of higher ice concentration. We formulate an objective function for the total collision cost based on a kinetic energy loss formulation of the ship-ice interaction. 
    
    \item An optimization-based improvement step to locally refine the solution of a lattice-based planner, reducing the navigation cost in a two-stage path planning pipeline

    \item An open source Python-based 2D simulator for efficient experimentation of full-scale ship transit in ice floe fields, addressing the current lack of any available open source physics simulators for ship operations in broken ice. 

    \item Results from extensive simulation experiments demonstrating our method's ability to significantly reduce both the mean and maximum impact forces from ship-ice collisions and total energy consumption compared to naive navigation strategies.

    \item Results from physical experiments conducted in a model-scale setup that show a 46\% improvement in performance over prior work. 

\end{enumerate}

\subsection{Related Work}
We review existing work relevant to our problem of interest, beginning with ship autonomy in ice navigation followed by related work on trajectory and motion planning in other maritime contexts. We note that while the problem of ice navigation shares similarities with a broader set of problems in robotics known as Navigation Among Movable Obstacles (NAMO) \cite{stilman_planning_2008}, the agent-obstacle interactions involve additional complexities (e.g., risk of damage to the agent and obstacles interacting with each other) when navigating among ice floes. 

\subsubsection{Ship autonomy for ice navigation}

Many studies have considered the task of optimizing a global route for ice navigation \cite{tran2023pathfinding, nam2013simulation, ryan2021arctic, choi2015arctic, Montewka2019, ZHANG2019106071, lehtola2019finding, kotovirta2009system, smith2022autonomous}. The goal in this task is to compute a sequence of waypoints from a starting position to a target destination that minimizes an objective function subject to a set of constraints. These routes are computed online and updated as necessary to provide ship captains with real-time guidance. However, these routes only provide navigational guidance at the global level and are optimized based on a coarse representation of the environment.  For example, in \cite{lehtola2019finding, kotovirta2009system, smith2022autonomous},  a discrete mesh is generated to represent each relevant environmental condition, with each cell in the mesh representing a subregion and storing the corresponding input data value (e.g., sea ice concentration or ice thickness).
Using ship performance models such as \cite{ryan2021arctic, colbourne2000scaling}, the mesh representation of the environment can be converted to a discrete map of navigation costs. Thus, the route planning problem can be efficiently solved by generating a graph from this discrete costmap. In our work, we follow a similar approach for representing the navigation cost, where we assign a collision cost to each cell in a fine grid representing the local ice conditions. Our cost function is also motivated by the objective functions typically considered in route planning but considers the cost at the level of individual ship-ice collisions along the path. Both the route planner and the local planner systems are required to achieve full ship autonomy \cite{bolbot2023small}.
% artifacts occur due to the coarse representation 
% mention speed limit given ice conditions
% ice resistance force and what it is 

Few studies consider the local planning task for ice navigation. In total, we found four previous works that tackle this problem \cite{aksakalli2017optimal, ari2013optimal, hsieh2021sea, gash2020machine}. In \cite{aksakalli2017optimal, ari2013optimal, hsieh2021sea}, the authors propose a method for computing collision-free paths in broken ice environments. Although the authors of \cite{aksakalli2017optimal, ari2013optimal} consider the turning radius constraints of the ship, the collision-free assumption limits the practical relevance of these works in ice navigation. Recall that a key assumption in our work is that the ice environment typically precludes collision-free navigation. The final local navigation approach we found is the shortest open water planning approach described in \cite{gash2020machine}. Here, the authors leverage the popular image processing technique, morphological skeletonization, to create a topological representation of the open water area from an overhead snapshot of the ice field. The information in the processed image is then extracted to build a graph from which a path is computed. Among these four related works \cite{aksakalli2017optimal, ari2013optimal, hsieh2021sea, gash2020machine}, the most practical method is \cite{gash2020machine} since it can easily be extended to high-concentration ice fields. As a result, we considered this approach as one of the baselines in our experiments.

\subsubsection{General ship autonomy}
A larger body of work considers the broader task of autonomous ship operation in non-ice environments \cite{chiang2018colreg, zhuang2011motion, bergman2020optimization, bergman2019improved, bergman2020, shan2020receding, bitar2018energy, ruud2023hybrid}. This includes urban waterways \cite{shan2020receding}, harbors \cite{bitar2020two, bergman2020}, and shorelines \cite{chiang2018colreg}. 
In \cite{shan2020receding}, a receding-horizon path planner is proposed for an autonomous ship operating in constrained urban environments. Since the obstacle environment is dynamic in nature, local planning is performed in an iterative fashion, where paths are continuously generated in real-time up to a small distance ahead of the ship's current position. These paths should therefore account for the kinodynamic constraints of the vessel to be effectively followed. A challenge here is determining how to efficiently represent the space of feasible paths or, in the case of motion planning, the space of possible maneuvers that can be executed. Many methods address this problem by considering a discretization scheme. One popular technique is to define offline a finite set of optimized maneuvers as introduced in \cite{pivtoraiko2009differentially}. These maneuvers can be repeated in a particular sequence to produce a planned trajectory. The local planning problem can therefore be reduced to finding a cost-minimizing sequence of these maneuvers using graph search techniques \cite{lavalle2006planning}. In this work, we adopt the so-called state lattice planning approach to produce reference paths that consider the minimum turning radius constraints of the ship.

Treating local planning as a graph search problem has been shown to be an effective technique for autonomous ship navigation. In \cite{bergman2020}, the authors develop a two-stage optimization-based motion planner for ships navigating in confined environments (e.g., harbors). In the first stage, a lattice planner computes a trajectory by searching over a graph of motion primitives, while in the second stage, the authors formulate an optimal control problem which is warm-started with the lattice planning solution. This two-stage approach ensures that the computed plan is locally optimal with respect to the continuous planning problem. A similar method is proposed in \cite{bitar2020two}, a key contribution being their novel encoding of polygonal obstacles into smooth and convex constraints. These works motivate our proposal for an optimization-based improvement step to enhance the path computed by the lattice planning stage in our ice navigation framework. We place particular emphasis on how we define a smooth objective function from the discrete collision cost formulated as part of the lattice planner stage.

\section{Preliminaries}
\subsection{Notation and Ship Modeling}
\label{sec:prelim_notation}
\begin{figure}[t]
    \centering
    \includegraphics[width=\columnwidth]{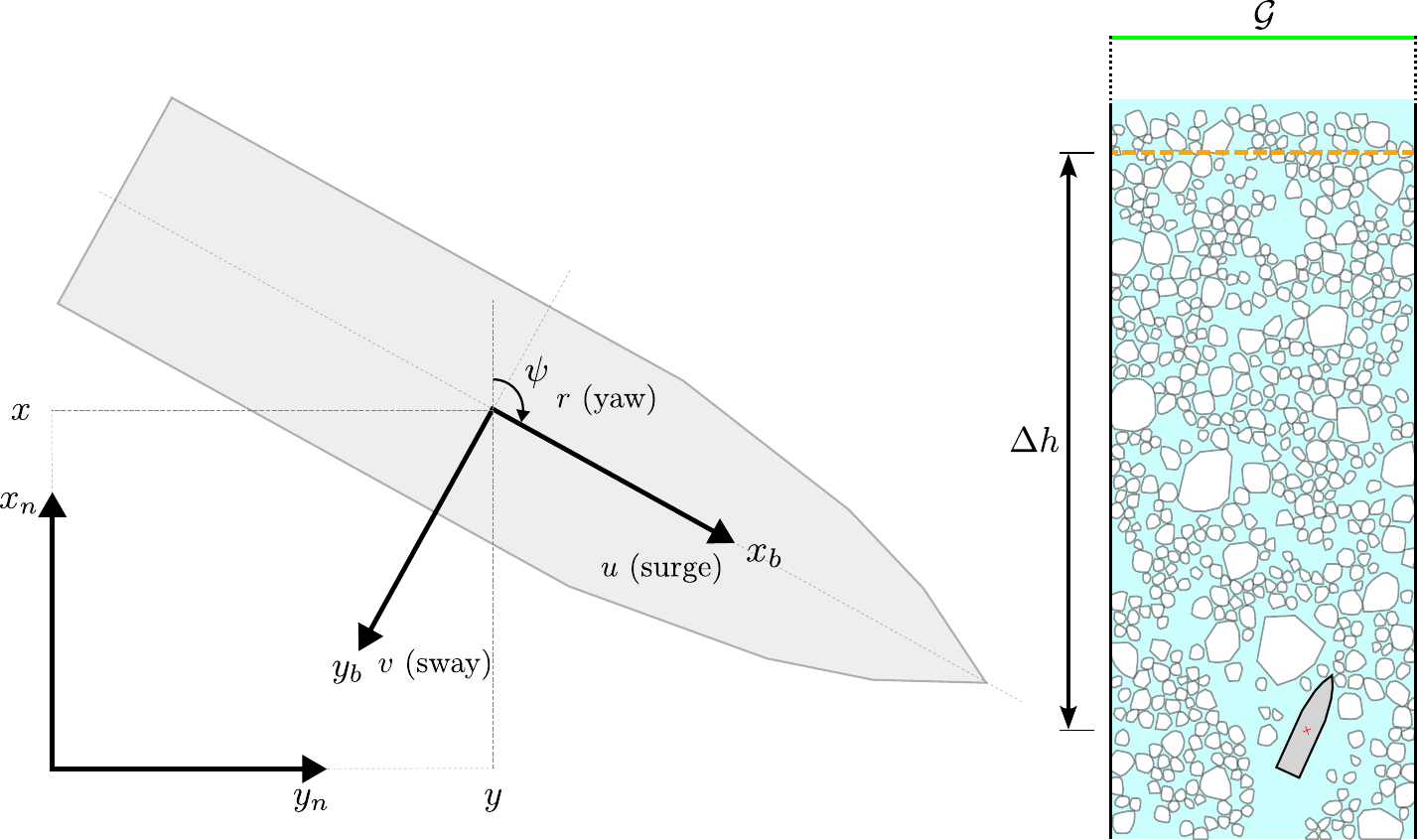}
    \caption{(Left) The two coordinate systems used to describe ship motion in three degrees of freedom. (Right) Depiction of the navigation problem of interest for a ship navigating a cluttered ice channel where the high-level objective is forward progress along the channel.}
    \label{fig:prob_formulation} 
\end{figure}

In this section, we introduce the notation used in this paper following the standard notation for the motion of a marine craft described in \cite{fossen2011handbook, sname1950nomenclature}. The ocean surface on which a ship moves is treated as a 2D surface $\mathcal{W} \subseteq \mathbb{R}^2$. As such, it is sufficient to consider 3 degrees of freedom (DoF) for ship motion in the horizontal plane \cite{fossen2011handbook}. This corresponds to two translational movements, \emph{surge} and \emph{sway}, and one rotational movement, \emph{yaw}, as shown in Fig.~\ref{fig:prob_formulation} (left). We use two reference frames to characterize the motion of the ship: an inertial frame $(x_n, y_n)$ and a body-fixed reference frame $(x_b, y_b)$ with the origin $o_b$ located at the center of gravity of the ship. The coordinates of $o_b$ expressed in the inertial frame give the position of the ship $(x, y) \in \mathbb{R}^2$. The direction in which the front of the ship points is referred to as the heading angle $\psi \in [0, 2 \pi) = \mathbb{S}^1$ which is defined by the angle from the inertial axis $x_n$ to the body-fixed axis $x_b$. Together, the position and heading are referred to as the generalized position or the \emph{pose} of the ship, denoted $\bm{\eta}$ where
\begin{equation}
    \bm{\eta} = \begin{bmatrix}x & y & \psi\end{bmatrix}^\top \in \mathcal{C} = \mathbb{R}^2 \times \mathbb{S}^1.
\end{equation}
The set $\mathcal{C}$ is considered the robot \emph{configuration space} \cite{lavalle2006planning} of the ship. The velocity of the ship is expressed in the body-fixed reference frame. The three components are surge velocity $u$, sway velocity $v$, and yaw rate $r$. As a vector, the generalized velocity is represented by
\begin{equation}
    \bm{\nu} = \begin{bmatrix}u & v & r\end{bmatrix}^\top \in \mathbb{R}^3.
\end{equation}
We can compute $\sqrt{u^2 + v^2}$ to obtain the speed $U \in \mathbb{R}$. The vectors $\bm{\eta}$, $\bm{\nu}$ are related via the kinematic relationship
\begin{equation}
    \dot{\bm{\eta}} = \mathbf{R}(\psi)\bm{\nu},
\end{equation}
where
\begin{equation}
    \mathbf{R}(\psi) =
\begin{bmatrix}
\cos\psi & -\sin\psi & 0 \\
\sin\psi & \cos\psi & 0 \\
0 & 0 & 1
\end{bmatrix}
\end{equation}
is the rotation matrix. A visual summary of the notation described above is provided in Fig.~\ref{fig:prob_formulation} (left). 

Standard ship terminology is referenced in this work. The watertight body of the ship is called the \emph{hull} and the front and rear parts are called the \emph{bow}, and \emph{stern}, respectively. The design of the hull varies greatly between different ship classes. In the case of ice-going vessels, the hull and other parts of the ship are reinforced to a certain level and are characterized by their assigned \emph{ice class} \cite{canada_2019}. Given particular ice conditions, the ice class determines a safe vessel speed under which the expected resistance forces remain within an acceptable level.

\subsection{Problem Formulation}
\label{sec:prob_formulation}

We consider the task of navigating a ship through a broken ice field, modeled as a rectangular ice channel $\mathcal{I} \subseteq \mathcal{W}$  as in Fig.~\ref{fig:prob_formulation} (right). Without loss of generality, the channel length is assumed to be aligned with the $x_n$ axis of the inertial frame. The high-level objective is for the ship to make forward progress along the ice channel. As such, we consider a line segment perpendicular to the channel length as the navigation goal $\G$, defined in the configuration space $\mathcal{C}$, as follows: 
\begin{equation}
    \G = \{ \bm{\eta} \in \mathcal{C} \mid (x, y) \in \mathcal{I}, x = x_{\text{goal}} \},
\label{eq:goal}
\end{equation}
with parameter $x_{\text{goal}}$ denoting the $x$ coordinate of the goal. Note that while we treat the navigation area as rectangular, it is possible to adapt our method to other environment shapes\footnote{One option is to partition the environment into a sequence of circumscribed rectangles and to assign an appropriate cost of navigating in the regions outside of the channel boundaries}.

Suppose a top-down view of the ice field is actively captured with sensors mounted on board \cite{zhang2022semantic} or with unmanned aerial vehicles \cite{zhang2014image} and processed via an ice image processing system \cite{zhang2018sea}. Given $m$ ice floes observed in the environment, each floe is treated as an obstacle with a \emph{footprint} $\mathcal{O}_i \subset \mathcal{I}$ occupying space in the 2D environment, where $i = 1, \dots, m$. Using the footprints, we can extract important ice properties, including shape and size \cite{zhang2022semantic}. Let $\obs=\{\mathcal{O}_1,\dots,\mathcal{O}_m\}$ denote the collection of all ice floes and let $\mathcal{W}_{\text{free}}$ denote the corresponding open water area defined as the set difference $\mathcal{I} \setminus \bigcup_{i} \mathcal{O}_i$. The notion of a footprint is also useful for characterizing the region in $\mathcal{I}$ occupied by the ship at a pose $\bm{\eta}$, denoted as $\mathcal{F}(\bm{\eta}) \subset \mathcal{I}$.

To navigate a ship, we require the following: a \emph{reference path} $\bm{\pi}: \mathbb{R} \rightarrow \mathcal{C}$ parameterized by arc length $s \in [0, L_f]$ with total path length $L_f$, a \emph{velocity profile} $\bm{\mathrm{v}}: \mathbb{R} \rightarrow \mathbb{R}^3$ that describes how the path is traversed over time, and a \emph{controller} that tracks the path at the specified velocity. Let the tuple $(\bm{\pi}, \bm{\mathrm{v}})$ denote a \emph{trajectory}. To capture the navigation cost of a trajectory $(\bm{\pi}, \bm{\mathrm{v}})$, we define an objective function $J$ as
\begin{equation}
       J = T_f + \alpha C_f,
\label{eq:prob_formulation_J_cost}
\end{equation}
which penalizes a weighted sum of the total travel time $T_f$ and the \emph{total collision cost} $C_f$, weighted by the parameter $\alpha \geq 0$.
Although the precise form of $C_f$ will be discussed later, we propose that a suitable function for the cost incurred from ship-ice collisions should accumulate a collision cost $c_{\text{obs}} \geq 0$ as the ship, characterized by its footprint $\mathcal{F}$, follows a trajectory $(\bm{\pi}, \bm{\mathrm{v}})$ through an ice channel containing ice floes $\obs$.
Thus, we seek to solve the following problem.
\begin{problem}[Ice Navigation Trajectory Planning Problem]
\label{prob:main_prob}
Given an initial pose $\bm{\eta}_{\textnormal{cur}}$, a goal $\G$, ice floes $\mathcal{W}_{\textnormal{obs}}$, a ship with a footprint $\mathcal{F}$, and an ice channel $\mathcal{I}$, compute a trajectory $(\bm{\pi}, \bm{\mathrm{v}})$ from $\bm{\eta}_{\textnormal{cur}}$ to $\G$ that minimizes $J$ and such that $\bm{\mathrm{v}}$ is bounded above by a safe vessel speed.
\end{problem}

In what follows, we propose a solution to Problem \ref{prob:main_prob} and describe how it is incorporated into a navigation system.

\section{Navigation Framework}
\label{sec:approach}
We present a navigation framework and an overview of our approach to solving Problem \ref{prob:main_prob}. To account for the large, evolving environment, we repeatedly replan a reference path and velocity profile for a controller in real time. Replanning occurs over a moving horizon (orange line in Fig.~\ref{fig:prob_formulation} (right)).

\begin{algorithm}
\caption{AUTO-IceNav}
 \hspace*{\algorithmicindent} \textbf{Input:} $\G, \Delta h, \Delta t$
\label{alg:GeneralPlanner}
\begin{algorithmic}[1]
\While{\texttt{True}}
\State $\bm{\eta}_{\textnormal{cur}} \gets $ Current ship pose
\If{$\bm{\eta}_{\textnormal{cur}}$ has not crossed $\G$}
\State Obtain $\obs$ and set subgoal $\G_{\text{int}}$ given $\Delta h$, $\bm{\eta}_{\textnormal{cur}}$
\State $U_{\text{nom}} \gets $ Nominal speed given $\obs$
\State $c_{\text{obs}}^{(1)} \gets$ KE cost given $\obs$, $U_{\text{nom}}$ \Comment{Sec~\ref{sec:ke_penalty}}
\State $c_{\text{obs}}^{(2)} \gets$ Ice coverage cost given $\obs$ \Comment{Sec~\ref{sec:con_penalty}}
\State $\mathcal{C}_{\text{map}} \gets$ Costmap given $c_{\text{obs}}^{(1)} c_{\text{obs}}^{(2)}$  \Comment{Eq. \ref{eq:collision_cost2} \quad \  \ }
\State $\bm{\pi} \gets $ Path $\bm{\eta}_{\textnormal{cur}} \leadsto \G_{\text{int}}$ given $\mathcal{C}_{\text{map}}$ \Comment{Sec \ref{sec:state_lattice} \quad}
\State $\bm{\pi}_{\text{opt}} \gets $ Optimized path given $\bm{\pi}$,  $\mathcal{C}_{\text{map}}$ \Comment{Sec \ref{sec:optim_step} \ \quad}
\State $\bm{\mathrm{v}}_{\text{nom}} \gets$ Velocity profile for $\bm{\pi}_{\text{opt}}$ given $U_{\text{nom}}$
\State Send $(\bm{\pi}_{\text{opt}}, \bm{\mathrm{v}}_{\text{nom}})$ to controller to track for time $\Delta t$
\Else
\State Exit loop
\EndIf
\EndWhile
\end{algorithmic}
\end{algorithm}

The navigation framework is summarized in Algorithm~\ref{alg:GeneralPlanner}, which takes as input a final goal $\G$, a receding horizon parameter $\Delta h$, and a tracking duration parameter $\Delta t$. At the start of each iteration, we obtain the current pose $\bm{\eta}_{\textnormal{cur}}$ and the updated ice floe information $\obs$ from the vision system (Lines 2, 4) --- we assume that other important ice properties may be queried at this stage, including the ice thickness and ice density, but it is often sufficient to treat these as constant for a given local region \cite{smith2022autonomous}. Using this new information, we update the horizon planning window and set a new limit $U_{\text{nom}}$ on the nominal safe speed (Lines 4, 5). The intermediate goal $\G_{\text{int}}$ is set at a distance $\Delta h$ ahead of the ship's position,  where $\Delta h$ is based on the vision system's effective perception range (typically 200–1000 meters for ship-mounted sensors \cite{zhang2018sea}).

The core of our proposed framework lies in Lines 6-10 of Algorithm~\ref{alg:GeneralPlanner}. Given the observed ice floes $\obs$ and the current nominal speed $U_{\text{nom}}$, we compute a costmap $\mathcal{C}_{\textup{map}}$ based on two different penalties associated with collisions involving the ice floes. In Line 6, we calculate a cost based on the ship's kinetic energy loss to penalize direct contact with ice floes, while in Line 7, we apply an additional penalty based on the ice concentration. Using a lattice-based planner, we can efficiently compute the total collision cost $C_f$ using the costmap $\mathcal{C}_{\textup{map}}$. In Line 9, the lattice planner plans a reference path $\bm{\pi}$ from the current ship pose $\bm{\pi}(0) = \bm{\eta}_{\textnormal{cur}}$ to somewhere along the intermediate goal $\bm{\pi}(L_f) \in \G_{\text{int}}$. From here, we locally optimize the planned path using our proposed optimization-based improvement step (Line 10). Since the path planning stage assumes that the ship is traveling at the nominal speed $U_{\text{nom}}$, we see that the travel time $T_f$ is equal to $L_f / U_{\text{nom}}$. Hence, our planning algorithm minimizes an objective that is equivalent to minimizing \eqref{eq:prob_formulation_J_cost}:
\begin{equation}
       J = L_f + \alpha C_f.
\label{eq:nav_framework_J_cost}
\end{equation}

In the final steps, the velocity profile $\bm{\mathrm{v}}_{\text{nom}}$ is generated for the optimized path (Line 11) and the controller tracks the planned trajectory $(\bm{\pi}_{\text{opt}},\bm{\mathrm{v}}_{\text{nom}})$ for time $\Delta t$ (Line 12). We discuss our implementation of these two final steps as part of the experimental setup in Section~\ref{sec:sim}.

The steps described above are repeated until the ship reaches the goal $\G$. Note that in each planning iteration, the ice is treated as static when computing a trajectory $(\bm{\pi}_{\text{opt}},\bm{\mathrm{v}}_{\text{nom}})$. However, we frequently perform planning updates to account for ship-ice collisions that cause changes in the ice environment over short periods of time. As a result, we maintain low runtime in our path planning algorithm. The following sections describe the two stages of our proposed path planning pipeline.

\section{Path Planning using a State Lattice}
\label{sec:state_lattice}
The lattice-based planner from \cite{pivtoraiko2009differentially} is adopted for our path planner, where planning is cast as a graph search problem. A finite set of precomputed actions, or motion primitives, is defined based on a robot's differential constraints. Repeated application of these primitives generates a lattice—a regularly spaced arrangement of reachable states in the robot's state space. Formally, each primitive $\bm{\pi}$ in a \emph{control set} $\mathcal{P}$ maps a  point $p_0 \in \mathcal{L}$ to another point $p_1 \in \mathcal{L}$ for a lattice $\mathcal{L}$. This defines a graph $G^{\mathcal{P}} = (\mathcal{L}, E, w)$, where edges $E$ represent feasible transitions generated by applying, rotating, and translating motion primitives, and $w$ assigns traversal costs. Standard graph search algorithms can then be used to efficiently plan cost-optimal feasible paths over $G^{\mathcal{P}}$. 

We discuss how the above-mentioned state lattice framework is applied to our path planning problem. Following several prior works on ship navigation on a 2D surface \cite{kim2014angular, ruud2023hybrid, caillau2019zermelo, liang2020path}, we assume paths generated by a ship can be approximated by a unicycle model where the curvature $\kappa$ is constrained to a limit $\kappa_{\max}$:
\begin{equation}
    \begin{split}
    \dot{\bm{\eta}}(s) &= \begin{bmatrix}\dot{x} \\ \dot{y} \\ \dot{\psi}\end{bmatrix} = \begin{bmatrix}
        \cos(\psi) \\ \sin(\psi) \\ \kappa
    \end{bmatrix}\\
        |\kappa| &\leq \kappa_{\max}.
    \end{split}
\label{eq:unicycle}
\end{equation}
Here, the derivatives $\dot{(\cdot)}$ are taken with respect to arc length $s$. We set the limit $\kappa_{\max}$ based on the physical limits of the ship given the particular conditions in the environment and the nominal speed $U_{\text{nom}}$ \cite{canada_2019, fossen2011handbook}. This limit is typically expressed in terms of the minimum turning radius of the ship, $r_{\min}$, where $r_{\min} = \kappa_{\max}^{-1}$.

Given $\eqref{eq:unicycle}$, we can consider the discretization of the configuration space $\mathcal{C}$. We generate a state lattice $\mathcal{L}$ by discretizing the plane $\mathbb{R}^2$ into a uniform grid and the unit circle $\mathbb{S}^1$ into uniformly spaced angles. In each iteration of Algorithm \ref{alg:GeneralPlanner}, we build a graph $G^{\mathcal{P}}$ with the lattice initialized at the current pose of the ship $\bm{\eta}_{\textnormal{cur}}$. To define our motion primitives, we compute shortest paths between states $\bm{\eta} = [x \ y \ \psi]^\top$ for the unicycle model \eqref{eq:unicycle}, known as Dubins' paths~\cite{dubins1957curves}. These paths consist of sequences of straight lines and circular arcs of radius $r_{\min}$. Ideally, we would like to have a minimum turning radius $r_{\min}$ that accurately reflects the current nominal speed $U_{\text{nom}}$ from Algorithm \ref{alg:GeneralPlanner}. However, since the motion primitives are defined offline, we can instead generate a suite of control sets with different minimum turning radii and then select online the best matching control set and $r_{\min}$ for the given $U_{\text{nom}}$. The control set $\mathcal{P}$ (or a series of control sets with different $r_{\min}$) is generated using the method proposed in \cite{botros2021multi}. The edges of the graph are assigned a cost equal to our objective function $\eqref{eq:nav_framework_J_cost}$ where the total collision cost $C_f$ is described in detail in the next section. We use A* search \cite{hart1968formal} to plan a path $\bm{\pi}$ from the current pose $\bm{\eta}_{\textnormal{cur}}$ to the intermediate goal $\G_{\text{int}}$. An admissible heuristic used to improve the performance of A* is described in Appendix \ref{app:appendix_A_heuristic}.

\subsection{Total Collision Cost}
\label{sec:cost_fn}

If we assume ice-free conditions, the path length $L_f$ provides an approximate model for the ship's energy consumption in a local region where other environment conditions are likely less variable. As a result, for the total collision cost $C_f$ we use the notion of total \emph{kinetic energy} loss during ship-ice collisions to capture the additional cost incurred during navigation. We start by describing how we can efficiently compute the cost of a candidate path, $\bm{\pi}$, given obstacles, $\obs$, in a lattice-based planner.

\subsubsection{Costmap representation of the environment}
The ice channel $\mathcal{I}$ is discretized into a uniform grid with resolution $\Delta_{\text{grid}}$ where each grid cell is assigned an identifying tuple $\bm{k} \in \mathbb{N}^2$ corresponding to the position of the grid cell's center. We use a subscript `d' to distinguish between the continuous and discrete representations. Let $\mathcal{I}_d \subseteq \mathbb{N}^2$ denote the discretized ice channel. Each ice floe is mapped to a discrete footprint $\mathcal{O} \subset \mathcal{I}_d$ and we denote $\mathcal{W}_{\text{obs}, d}$ as the set of obstacle discrete footprints. Each grid cell $\bm{k}$ is an element of at most one obstacle footprint $\mathcal{O} \in \mathcal{W}_{\text{obs}, d}$ and $\bm{k}$ is considered an element of an obstacle $\mathcal{O}$ if the continuous footprint representation covers any part of the cell. 

Using this discrete representation of the planar environment, we can generate a costmap $\mathcal{C}_{\textup{map}}: \mathcal{I}_d \rightarrow \mathbb{R}_{\geq 0}$ \cite{pivtoraiko2009differentially} by assigning a collision cost to each grid cell using our proposed collision cost function, $c_{\text{obs}}$. Given the nominal speed $U_{\text{nom}}$ and the obstacles $\mathcal{W}_{\text{obs}, d}$, the cost at a grid cell $\bm{k}$ is defined as

\begin{equation}
\label{eq:collision_cost2}
     c_{\text{obs}}(\bm{k},U_{\text{nom}},\mathcal{W}_{\text{obs}, d}) = 
     \begin{cases}
   c_{\text{obs}}^{(1)}c_{\text{obs}}^{(2)}, & \exists \mathcal{O} \mid \bm{k} \in \mathcal{O}\\
    0  &\text{otherwise}, 
    \end{cases},
\end{equation}
where $c_{\text{obs}}^{(1)}$ is the kinetic energy loss penalty (Section~\ref{sec:ke_penalty}) and $c_{\text{obs}}^{(2)}$ is the ice concentration penalty (Section~\ref{sec:con_penalty}). Observe that \eqref{eq:collision_cost2} results in a costmap with nonzero costs only at grid cells that overlap with an obstacle.

\begin{figure}[t]
    \centering
    \includegraphics[width=\columnwidth]{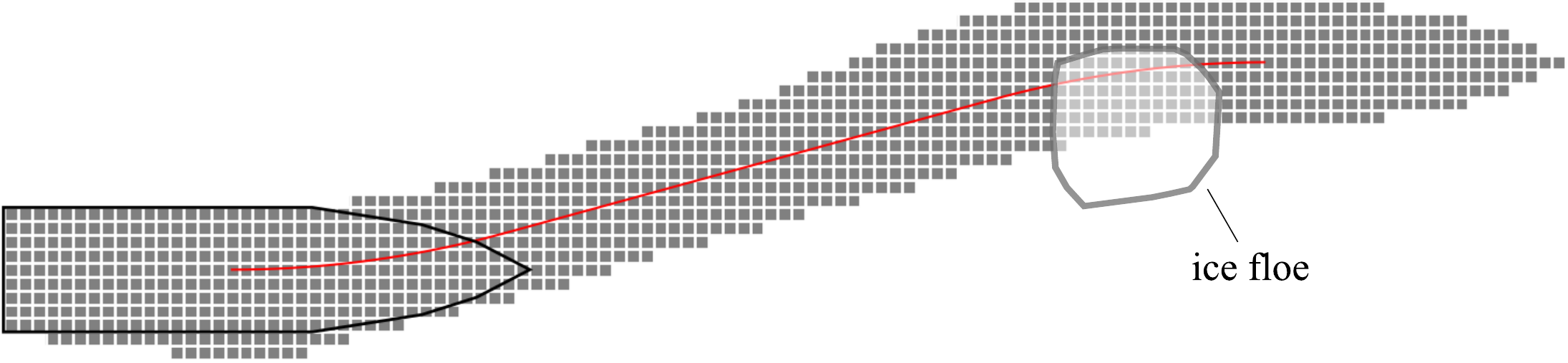}
    \caption{The swath (grey grid cells) of a sample path (red line) and ship footprint (black outline) overlapping one ice floe. In this example, each cell in the discretized channel represents a region of size 2 m $\times$ 2 m.}
    \label{fig:costmap_resolution}
\end{figure}

To compute the cost of being at the pose $\bm{\eta}$ we require an appropriate mapping from the configuration space to $\mathcal{I}_d$ as well as a footprint $\mathcal{F}_d(\bm{\eta}) \subset \mathbb{N}^2$ specifying the set of grid cells occupied by the ship body given the current position and heading. The total collision cost for a path $\bm{\pi}$ is computed as a sum over the set of costmap grid cells $\mathcal{S}(\bm{\pi}) \subset \mathcal{I}_d$ swept by the ship footprint $\mathcal{F}_d$ from tracing the path $\bm{\pi}$. The set $\mathcal{S}(\bm{\pi})$ is referred to as the \emph{swath} \cite{pivtoraiko2009differentially} of the path $\bm{\pi}$, and we show an example in Fig.~\ref{fig:costmap_resolution}. Hence, we compute the total collision cost $C_f$ as the swath cost\footnote{In addition to precomputing the motiom primitives, swaths are precomputed to enable efficient computation of the swath cost.}:
\begin{equation}
    C_f = \sum_{\bm{k} \in \mathcal{S}(\bm{\pi})} c_{\text{obs}}(\bm{k},U_{\text{nom}},\mathcal{W}_{\text{obs}, d}).
\label{eq:swath_cost}
\end{equation} 

\begin{remark}[Obstacle Proximity Penalty]
In contrast to typical cost functions used in navigation problems, \eqref{eq:swath_cost} does not apply a penalty for being in proximity to obstacles that are not in collision with the path. In our particular problem, planning paths arbitrarily close to obstacle edges (particularly large ice floes) without penalty may not be desirable behavior, especially considering that we expect some error in tracking. To address this, a straightforward extension is to scale up the obstacle footprints $\obs$ by a small factor (e.g., 0.1) proportional to their size. This would effectively generate a cost buffer around the ice floes in the costmap. Regarding the potential issue of having overlap between multiple scaled-up obstacle footprints, the solution used here is to assign the maximum of the costs computed for a grid cell $\bm{k}$.
\end{remark}

\subsubsection{Kinetic energy loss penalty}
\label{sec:ke_penalty}
We present the kinetic energy loss penalty $c_{\text{obs}}^{(1)}$ from \eqref{eq:collision_cost2}. We consider a simple 2D ship-ice collision model and show a derivation of the kinetic energy loss of the ship from the collision.
More complex ship-ice interactions, such as floe-splitting and rafting behavior \cite{canada_2020}, are ignored.

\begin{figure}[!t]
    \centering
    \includegraphics[width=\columnwidth]{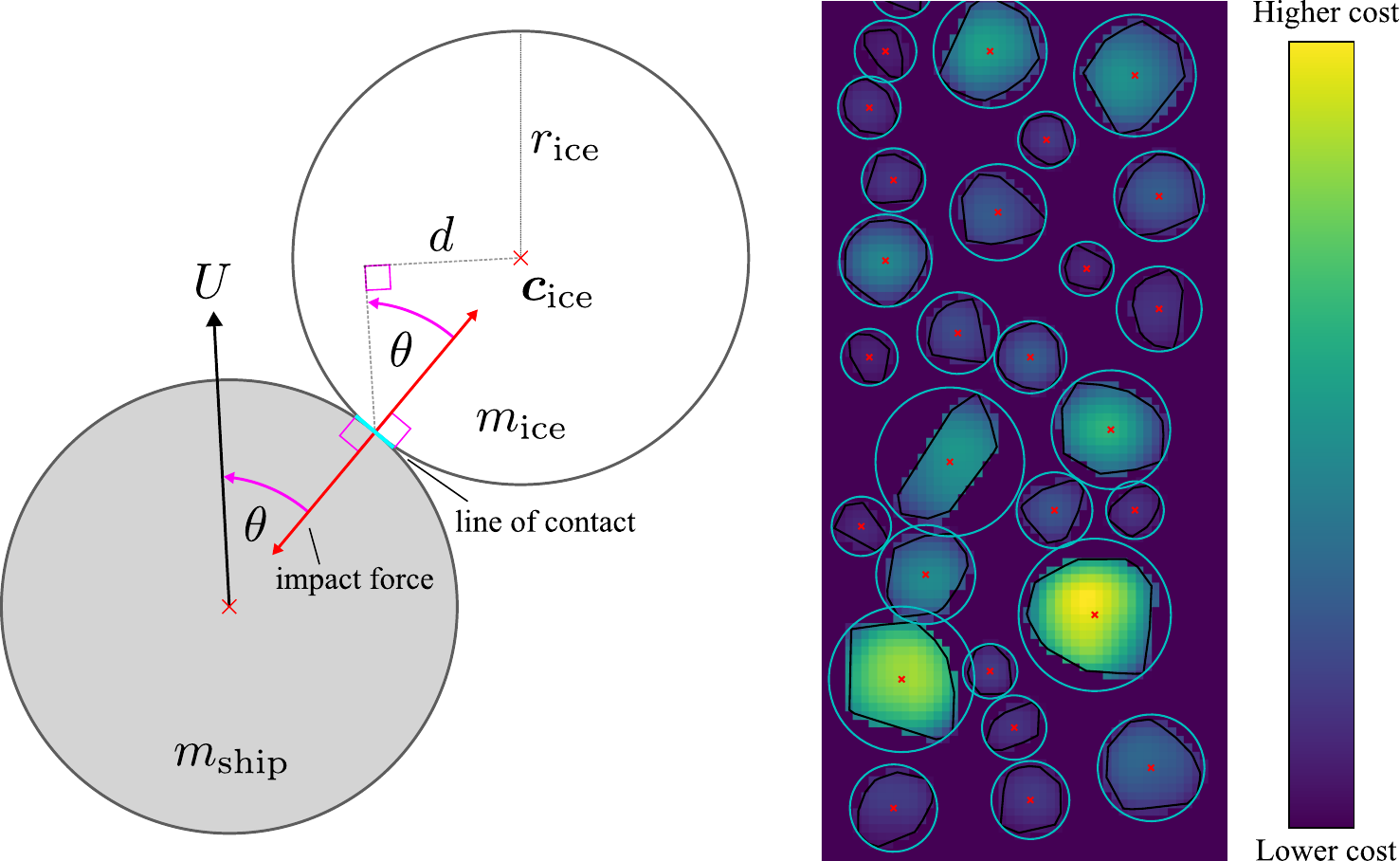} 
    \caption{(Left) The collision model of two disk-shaped bodies used to derive the cost function $c_{\text{obs}}^{(1)}$. (Right) An example of a costmap with a grid cell size of 1 m $\times$ 1 m. Brighter colors indicate higher cost. The centroid of each ice floe is indicated by the red `x' and the bounding circles are shown in light blue. In this example, the ice thickness and density is the same across ice floes meaning the size of the polygon corresponds to its mass.}
    \label{fig:collision}
\end{figure}

Let $m_{\text{ship}}$ and $m_{\text{ice}}$ denote the mass of the ship and ice floe, respectively. We assume that ice floes have uniform density and thickness. Thus, the mass $m_{\text{ice}}$ is given by the product of density, thickness, and area of the ice floe \cite{daley2014gpu}, where the area is determined by the extent of its footprint $\mathcal{O} \in \obs$ (usually a convex polygon). Similar to the collision model proposed in \cite{daley2014gpu} and then experimentally validated in \cite{alawneh2015hyper, kim2019evaluation}, we reduce the collision to a one-body problem by making the following assumptions: the collision is of short duration, the momentum of the system is conserved (i.e., friction is ignored), and the impact force is normal to the line of contact between the two bodies (see Fig.~\ref{fig:collision} (left)). In addition, the collision is assumed to be inelastic, where the maximum amount of kinetic energy of the system is lost to ice crushing \cite{daley1999energy, daley2014gpu}.

With this collision model, the change in kinetic energy $\Delta K_{\text{sys}}$ in the system is given by
\begin{equation}
\label{eq:delta_Ksys1}
    \Delta K_{\text{sys}} = -\frac{1}{2}M_{\text{eq}}V_{\text{eq}}^2,
\end{equation}
where $M_{\text{eq}}$, called effective mass, is the mass in the center of mass frame, and $V_{\text{eq}}$ is the relative velocity of the two bodies before the collision with respect to the line of contact. If we consider two disk-shaped bodies, the effective mass is only a function of the two masses:
\begin{equation}
    M_{\text{eq}} = \frac{m_{\text{ship}}m_{\text{ice}}}{m_{\text{ship}} + m_{\text{ice}}}.
\end{equation}
From \cite{daley1999energy}, the ice velocity is assumed to be negligible compared to the ship velocity. Hence, the relative velocity $V_{\text{eq}}$ is a function of the ship speed $U$ and the angle $\theta$ between the direction in which $U$ is pointing and the vector normal to the line of contact:
\begin{equation}
    V_{\text{eq}} = U\cos(\theta).
\end{equation}
The angle $\theta$ is illustrated in Fig.~\ref{fig:collision} (left). Next, we isolate the change in kinetic energy of the ship $\Delta K_{\text{ship}}$.

The goal of our collision cost is to minimize the kinetic energy loss of the ship due to the energy transferred to the ice and absorbed by ice crushing. Since the ice is treated as static prior to collision, the change in kinetic energy of the ice $\Delta K_{\text{ice}}$ is greater than 0. In summary, we have the inequalities
\begin{equation}
        \Delta K_{\text{sys}} < 0, \quad \Delta K_{\text{ice}} > 0,  \quad \Delta K_{\text{ship}} < 0,
\end{equation}
as well as the following relationship between the changes in kinetic energy for the ship, ice, and system:
\begin{equation}
\label{eq:delta_Ksys2}
    \Delta K_{\text{sys}} = \Delta K_{\text{ice}} + \Delta K_{\text{ship}}.
\end{equation}
After the collision has occurred, we treat the ice floe as having a final velocity equal to $V_{\text{eq}}m_{\text{ship}}(m_{\text{ship}} + m_{\text{ice}})^{-1}$. 
Therefore,
\begin{equation}
\begin{split}
    \Delta K_{\text{ice}} &= \frac{1}{2}m_{\text{ice}}\left(\frac{V_{\text{eq}}m_{\text{ship}}}{m_{\text{ship}} + m_{\text{ice}}}\right)^2\\
    &= \frac{1}{2}\frac{m_{\text{ice}}m_{\text{ship}}^2}{(m_{\text{ship}} + m_{\text{ice}})^2} (U\cos(\theta))^2
\end{split}
\end{equation}
and from \eqref{eq:delta_Ksys1} \eqref{eq:delta_Ksys2}\footnote{The expression given in our preliminary work \cite{deschaetzen2023} contained a small error.}:
\begin{equation}
\label{eq:delta_Kship1}
    \Delta K_{\text{ship}} = -\frac{m_{\text{ice}}m_{\text{ship}}(m_{\text{ice}} + 2m_{\text{ship}})}{2(m_{\text{ice}} + m_{\text{ship}})^2}(U \cos(\theta))^2.
\end{equation}

Suppose the disk representing the ice floe has a radius of $r_{\text{ice}}$ and a center located at $\bm{c}_{\text{ice}} \in \mathbb{R}^2$. The angle $\theta$ in the expression \eqref{eq:delta_Kship1} can be written in terms of $r_{\text{ice}}$ and the lateral distance $d$ between $\bm{c}_{\text{ice}}$ and the center of the line of contact (where impact force is exerted) measured with respect to the component parallel to the direction of $U$. From Fig.~\ref{fig:collision} we have,
\begin{equation}
    \theta = \arcsin \left(\frac{d}{r_{\text{ice}}}\right).
\end{equation}
As a result, given a particular ship with constant mass $m_{\text{ship}}$, $\Delta K_{\text{ship}}$ from \eqref{eq:delta_Kship1} can be written as a function of $d$, $r_{\text{ice}}$, $m_{\text{ice}}$, and $U$:
\begin{equation}
\begin{split}
       &\Delta K_{\text{ship}}(d, r_{\text{ice}}, m_{\text{ice}}, U) =\\
       &=\frac{m_{\text{ice}}m_{\text{ship}}(m_{\text{ice}} + 2m_{\text{ship}})}{2(m_{\text{ice}} + m_{\text{ship}})^2}\left [U \cos\left(\arcsin\left(\frac{d}{r_{\text{ice}}}\right)\right) \right ]^2\\
        & = \frac{U^2 m_{\text{ice}}m_{\text{ship}}(m_{\text{ice}} + 2m_{\text{ship}})}{2(m_{\text{ice}} + m_{\text{ship}})^2}\left(\frac{r_{\text{ice}}^2-d^2}{r_{\text{ice}}^2}\right), \quad   d \in [0, r_{\text{ice}}].
        \label{eq:delta_Kship2}
\end{split}
\end{equation}
Here, we have omitted the negative sign so that now $\Delta K_{\text{ship}}$ denotes the kinetic energy loss of the ship. 
Finally, consider a cell $\bm{k}$ and assume there exists an ice floe $\mathcal{O}_j\in\mathcal{W}_{\text{obs}, d}$ such that $\bm{k} \in \mathcal{O}_j$. Let $\bm{c}_{\text{ice},j}$ denote the position of the centroid of $\mathcal{O}_j$, $r_{\text{ice},j}$ the radius of its bounding circle centered at $\bm{c}_{\text{ice},j}$, and $m_{\text{ice},j}$ its mass. The kinetic energy loss penalty for cell $\bm{k}$ given current nominal speed $U_{\text{nom}}$ is defined as 
\begin{equation}
\label{eq:ke_penalty}
\begin{split}
    & c_{\text{obs}}^{(1)} = \Delta K_{\text{ship}}(d', r_{\text{ice}, j}, m_{\text{ice}, j}, U_{\text{nom}}),
\end{split}
\end{equation}
where $d'=||\bm{k}-\bm{c}_{\text{ice},j}||$ is the Euclidean distance from the grid cell $\bm{k}$ to the centroid $\bm{c}_{\text{ice},j}$ of the obstacle $\mathcal{O}_j$. Fig.~\ref{fig:collision} (right) illustrates a sample costmap and shows the bounding circles and radii of the obstacles.

\begin{observation}[Large Obstacles and Head-On Collisions]
Given the swath $\mathcal{S}(\bm{\pi})$ of a path $\bm{\pi}$ (e.g. Fig.~\ref{fig:costmap_resolution}), clearly from \eqref{eq:delta_Kship2}, \eqref{eq:ke_penalty} a larger penalty is applied if the swath overlaps ice floes with larger mass. In addition, the cost of the swath increases as the grid cells $\bm{k} \in \mathcal{S}(\bm{\pi})$ become closer to the centroid of an obstacle. This is particularly visible in Fig.~\ref{fig:collision} in the ice floes that are less circular (e.g., rectangular shaped), where the edges of the shape that are closer to the centroid are assigned a higher cost than the edges farther away. In practice, this means that paths with fewer head-on collisions with large ice floes are favored during planning. This typically results in the ship pushing the ice floes off to the side during navigation with few instances of sustained contact. Collisions with larger ice floes and collisions that are more head-on result in greater impact forces exerted on the ship \cite{canada_2019, daley2014gpu}. 
\end{observation}

\subsubsection{Ice concentration penalty}
\label{sec:con_penalty}
In the ice concentration penalty $c_{\text{obs}}^{(2)}$ from \eqref{eq:collision_cost2}, our goal is to account for secondary and higher-order collisions---scenarios where the ship displaces a greater amount of ice than the set of ice floes that were in direct contact with the ship. More formally, for a given path $\bm{\pi}$ and a corresponding swath $\mathcal{S}(\bm{\pi})$, there may exist a set of obstacles $\{\mathcal{O}_j\}_{j=1}^n \subset \mathcal{W}_{\text{obs}, d}$, where $n > 0$, and where for each $\mathcal{O}_j$ we have $\mathcal{O}_j \cap \mathcal{S}(\bm{\pi}) = \emptyset$, and yet each $\mathcal{O}_j$ creates additional resistance to ship motion, resulting in an increase in kinetic energy loss. Such scenarios are more likely to occur as the concentration of the ice field increases. Moreover, in \cite{colbourne2000scaling} the relationship between the average ice resistance force exerted on the ship and the ice concentration $C_{\text{ice}} \in [0, 1]$ is modeled as a linear function in $(C_{\text{ice}})^{\beta}$ where $\beta \geq 1$ is a constant parameter determined by the geometry of the particular ship hull. 

To assign an ice concentration cost to grid cells in the costmap, we first generate a binary occupancy image of the ice channel from $\mathcal{W}_{\text{obs}, d}$ indicating the pixels that are occupied by an ice floe. We then compute an image convolution between the binary image and a mean filter with kernel size $z \times z$, where $z$ is odd. For a given pixel $\bm{k} \in \mathbb{N}^2$ in the binary image, the mean filter will compute the average ice concentration for a region of size $z \times z$ in pixel units centered around $\bm{k}$. For the purpose of local navigation, the kernel should therefore be appropriately sized so that at a given location the extent of the kernel can capture a collection of ice floes in a region that is still relatively local compared to the rest of the ice channel. The resulting image, $C_{\text{ice}}: \mathbb{N}^2 \rightarrow [0, 1]$, provides a smooth representation of the ice concentration at every position in the discretized ice channel $\mathcal{I}_d$. This contrasts with the coarse grid or mesh representation \cite{smith2022autonomous}, where each cell represents a large area and is assigned a single ice concentration value, which is more suitable for route planning.

Note that we apply mirror padding to the binary image prior to computing $C_{\text{ice}}$ to prevent the loss of ice information at the boundaries of the image. Given the obstacles $\mathcal{W}_{\text{obs}, d}$, the ice concentration penalty for a cell $\bm{k}$ is defined as 
\begin{equation}
\label{eq:concentration_penalty}
    c_{\text{obs}}^{(2)} = (C_{\text{ice}}(\bm{k}))^{\beta}.
\end{equation}

\section{Optimization-Based Improvement Step}
\label{sec:optim_step}
In the previous section, we described our proposed lattice-based planner to efficiently compute low-cost reference paths for a tracking controller. While the planner computes optimal solutions to the discretized planning problem defined by the graph $G^{\mathcal{P}}$, planned paths often contain excessive oscillations \cite{botros2023spatio, dolgov2010path}, and the quality of the solution varies depending on how the lattice is initialized given the current pose of the ship $\bm{\eta}_{\textnormal{cur}}$. More broadly, the solutions are suboptimal for the following continuous path planning problem:

\begin{mini!}|l|[2]               
    {\bm{\pi}, \kappa, L_f}
    {J = L_f + \alpha C_f }
    {\label{eq:optim_continuous}} 
    {}                                % optimization result
    \addConstraint{\dot{x}(s) = \cos(\psi) \quad\quad &   s \in [0, L_f] }  \label{eq:constraint1}
    \addConstraint{\dot{y}(s) = \sin(\psi) \quad\quad &   s \in [0, L_f]}  \label{eq:constraint2}
    \addConstraint{\dot{\psi}(s) = \kappa(s) \quad\quad &   s \in [0, L_f]}  \label{eq:constraint3}
    \addConstraint{-r_{\min}^{-1} \leq \kappa(s) \leq r_{\min}^{-1} \quad\quad &   s \in [0, L_f]}  \label{eq:constraint4}
    % \addConstraint{(x(s), y(s)) \in \mathcal{I} \quad\quad &   s \in [0, L_f]}  \label{eq:con_uni5}
    \addConstraint{\mathcal{F}(\bm{\pi}(s)) \subset \mathcal{I} \quad\quad &   s \in [0, L_f]}  \label{eq:constraint5}
    \addConstraint{\bm{\pi}(0) = \bm{\eta}_{\textnormal{cur}}} 
    \addConstraint{\bm{\pi}(L_f) \in \G_{\text{int}},}
\end{mini!}
where the constraints \eqref{eq:constraint1}-\eqref{eq:constraint4} are from the unicycle model \eqref{eq:unicycle} and the constraint \eqref{eq:constraint5} ensures the ship body (characterized by the footprint $\mathcal{F}$) remains within the ice channel. In this version of the problem, the total collision cost $C_f$ extends the function \eqref{eq:swath_cost} to its continuous form, where we have an arc length parameterized line integral introduced in the next section. We will show how we can define a suitable differentiable function for the total collision cost that maintains the notions of ship footprint and swath using ideas from \cite{zucker2013chomp}.
Similar to \cite{dolgov2008practical, bergman2020, bitar2020two}, we propose an optimization-based improvement step in a two-stage planning pipeline. In particular, we solve the path planning problem \eqref{eq:optim_continuous} in the continuous domain producing a locally optimal solution to refine the path computed by the lattice planner, which is used as an initial feasible solution to warm-start an optimization solver.

In the first step, we define a smooth function $c_{\text{obs}}(x,y) \geq 0$ representing a scalar field for the collision cost from the grid-based costmap generated by the original discrete collision cost in \eqref{eq:collision_cost2}. We use bicubic interpolation to fit the costmap data to ensure that we have a smooth  function for the collision cost everywhere in the environment. This includes the boundaries of the obstacles where there are clear discontinuities in the discrete costmap.

\subsection{Continuous Function for Total Collision Cost}
With $c_{\text{obs}}(x,y)$ defined, we now need a way to calculate the total collision cost where there is a clear relationship with \eqref{eq:swath_cost}. We use the notion of body points from \cite{zucker2013chomp} which we define formally as follows:

\begin{definition}[Body Points]
Let $\mathcal{B} \subset \mathbb{R}^2$ denote a set of \emph{body points} where each body point $\bm{b} \in \mathcal{B}$ is a position $(x_{\mathcal{B}}, y_{\mathcal{B}})$ in the body-fixed coordinate frame (see Section \ref{sec:prelim_notation}).
\end{definition}

Consider the path of a particular body point $\bm{b}$ given a path $\bm{\pi}$. This gives the arc length parameterized curve $\bm{\pi}_{\bm{b}} = \{\bm{g}(\bm{\pi}(s), \bm{b}) : s \in [0, L_f]\},$ where the affine function $\bm{g}: \mathcal{C} \times \mathcal{B} \rightarrow \mathbb{R}^2$ maps a pose $\bm{\eta}$ and a body point $\bm{b}$ to the corresponding position in the world coordinate frame:
\begin{equation}
\label{eq:definition_g}
        \bm{g}(\bm{\eta}, \bm{b}) = \begin{bmatrix}
        \cos (\psi) & -\sin(\psi) \\ \sin(\psi) & \cos(\psi)
    \end{bmatrix}\bm{b} + \begin{bmatrix} x \\ y\end{bmatrix}.
\end{equation}
Next, consider the cost that the body point $\bm{b}$ collects in the cost field $c_{\text{obs}}(x,y)$  given by the line integral along the path $\bm{\pi}_{\bm{b}}$. We can therefore define the total collision cost $C_f$ as the cost of each of the body point paths integrated over all the body points:
\begin{equation}
\label{eq:optim_step_total_collision_cost}
    C_f = \int_0^{L_f} \int_{\mathcal{B}} c_{\text{obs}}(\bm{g}(\bm{\pi}(s), \bm{b})))\left \lVert \frac{d}{ds}\bm{g}(\bm{\pi}(s), \bm{b}))\right \rVert d\bm{b} \ ds.
\end{equation}
The operator $||\cdot||$ denotes the Euclidean norm of the vector $d\bm{g}/ds$, i.e., the arc length derivative of $\bm{g}$, which we can immediately get by looking at $\eqref{eq:definition_g}$ and recalling the unicycle model in the constraints \eqref{eq:constraint1}-\eqref{eq:constraint3}:
\begin{equation}
\label{eq:dg/ds}
\frac{d}{ds}\bm{g}(\bm{\eta}, \bm{b}) = \begin{bmatrix}
        -\kappa\sin (\psi) & -\kappa\cos(\psi) \\ \kappa\cos(\psi) & -\kappa\sin(\psi)
    \end{bmatrix}\bm{b} + \begin{bmatrix} \cos(\psi) \\ \sin(\psi)\end{bmatrix}.
\end{equation}
Note that in \eqref{eq:dg/ds}, we have suppressed the arc length parameter $s$ on the right-hand side of the equation; hence, $\kappa$ and $\psi$ should be interpreted as $\kappa(s)$ and $\psi(s)$, respectively. Multiplying the cost $c_{\text{obs}}$ by $\lVert d\bm{g} /ds \rVert$ ensures that we are integrating along the particular path $\bm{\pi}_{\bm{b}}$ for each body point $\bm{b} \in \mathcal{B}$, and not the path $\bm{\pi}$ unless $\bm{b} = [0 \ 0]^\top$.

\subsection{Options for an Appropriate Set of Body Points}
\begin{figure}[t]
    \centering
    \includegraphics[width=\columnwidth]{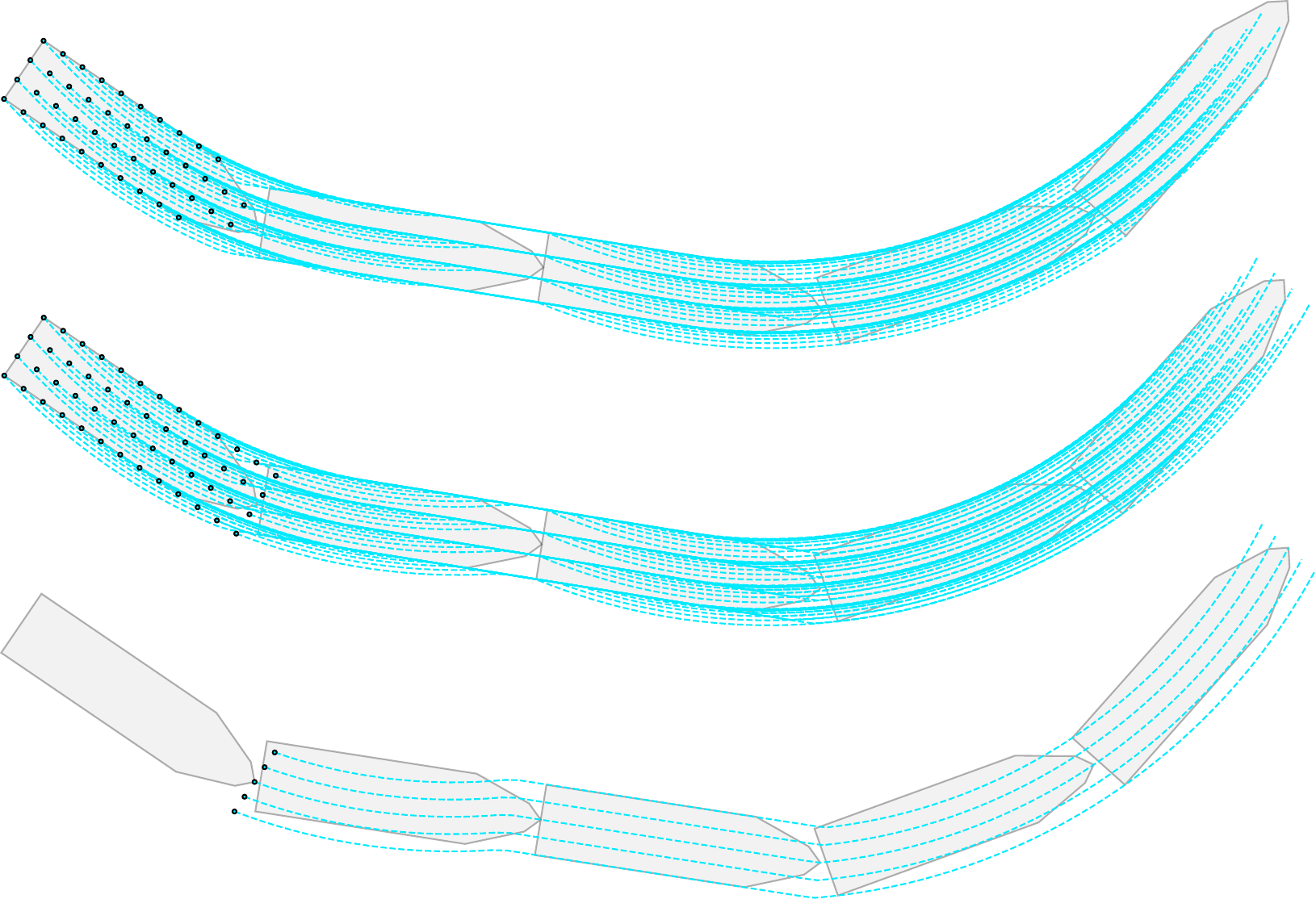}
    \caption{Three possible options to define the set $\mathcal{B}$. The dots are body points sampled from the respective $\mathcal{B}$ and the cyan dotted lines are the sampled body point paths.}
    \label{fig:body_points_options}
\end{figure}

We present three possible configurations for the set of body points $\mathcal{B}$, with an illustrative example of each shown in Fig.~\ref{fig:body_points_options}.
\begin{enumerate}[label=(\roman*)]
    \item Let $\mathcal{B}$ be equal to the ship footprint $\mathcal{F}(\bm{\eta})$ at $\bm{\eta} = [0 \ 0 \ 0]^\top$ (Fig.~\ref{fig:body_points_options} (top)). 
    \item Let $\mathcal{B}$ be the set of points inside the rectangle that encloses $\mathcal{F}(\bm{\eta})$ at $\bm{\eta} = [0 \ 0 \ 0]^\top$ (Fig.~\ref{fig:body_points_options} (middle)).
    \item Let $\mathcal{B}$ be the set of points along a straight line that is perpendicular to $\psi$, has a length equal to the width of the ship, and is positioned at the tip of the bow (Fig.~\ref{fig:body_points_options} (bottom)).
\end{enumerate}

\begin{remark}[Comparison with the Swath]
Consider how the set of body point paths from each of the three options for defining $\mathcal{B}$ compares to the swath $\mathcal{S}(\bm{\pi})$ given a path $\bm{\pi}$. Option (i) ensures no region from the swath is missed, option (ii) over-approximates the swath during turns, while option (iii) results in an under-approximation (albeit at a lower computational cost).
\end{remark}

\begin{remark}[Weighting Body Points]
    To enable fine-tuning of planning behavior, such as preferring collisions with areas of the hull that are reinforced, we can define a set of weights. Let each body point $\bm{b} \in \mathcal{B}$ be assigned a weight $w \geq 0$ representing the relative penalty for accumulating cost.
\end{remark}

Appendix \ref{app:body_points} provides a discussion comparing our choice of body points to the approach in \cite{zucker2013chomp}, and explains why, in our problem, defining body points exclusively along the ship’s outline would be incorrect.

\subsection{Transcription and Solver}
The continuous planning problem in \eqref{eq:optim_continuous}, incorporated with the collision cost $C_f$ from \eqref{eq:optim_step_total_collision_cost}, is solved numerically using tools from optimal control. In particular, we transform the continuous problem into a discrete nonlinear program (NLP) using direct multiple shooting. The path is discretized into $N+1$ decision variables $\bm{\eta}_1, \dots, \bm{\eta}_{N+1}$ and $N$ decision variables $\kappa_1, \dots, \kappa_N$ for the curvature $\kappa$ (i.e. the signal that controls $\dot{\psi}$). Each control interval occurs over an arc length step of $\Delta s$, analogous to a time step in trajectory optimization. To avoid large changes in curvature over a small distance, we add a smoothness term to the objective function with a weight parameter $\lambda \geq 0$. Integration of the ODE for the unicycle model is performed using the 4th-order Runge-Kutta method (RK4). 

\begin{figure}[t]
    \centering
    \includegraphics[width=\columnwidth]{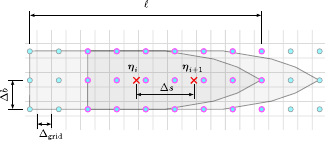}
    \caption{Illustration of the components involved to derive \eqref{eq:default_body_point_weight}. The square grid shows the discrete costmap with resolution $\Delta_{\text{grid}}$. Body points are shown for two consecutive poses, $\bm{\eta}_i, \bm{\eta}_{i+1}$, where the points outlined in magenta are overlapping. For the sake of visualization, everything here is nicely aligned with the costmap grid.}
    \label{fig:optim_weight}
\end{figure}

We define a finite set $\mathcal{B}_d$ of discrete body points uniformly spaced on a square grid with spacing $\Delta b$, covering the rectangle that encloses the ship body (i.e., option (ii) discussed in the previous section). Each body point $\bm{b}_j \in \mathcal{B}_d$ has a corresponding weight $w_j > 0$. Let $\ell$ denote the length of the ship body and recall that $\Delta_{\text{grid}}$ is the resolution of the discrete costmap. By default, each weight $w_j$ is assigned a constant value of 
\begin{equation}
\label{eq:default_body_point_weight}
    w_j = \frac{\alpha \Delta b^2}{\ell \Delta_{\text{grid}}^2},
\end{equation}
such that the terms in the objective---path length $L_f$ and total collision cost $C_f$---in the optimization problem have the same scaling as in the lattice planning stage. 
This expression is determined by calculating the average number of times a grid cell $\bm{k} \in \mathbb{N}$ in the original discrete costmap is covered by a body point $\bm{b}_j \in \mathcal{B}_d$ along a path $\{\bm{\eta}_i\}_{i=1}^{N+1}$ with an arc length step of $\Delta s$. Suppose we have a straight path (i.e. $\kappa_i = 0$ for $i=1,\dots,N$) where $\ell$, $\Delta s$, and $\Delta b$ are all multiples of $\Delta_{\text{grid}}$. A subset of this path is shown in Fig.~\ref{fig:optim_weight}. In this example, it is easy to see the two important ratios, namely $\Delta s / \ell $ and $\Delta b^2 / \Delta_{\text{grid}}^2$ which control how often a grid cell is sampled on average. These two ratios, along with the weight $\alpha$ from \eqref{eq:optim_continuous}, give \eqref{eq:default_body_point_weight} ($\Delta s$ is missing here since it appears in the objective function described next).

We remove the constraint $\eqref{eq:constraint5}$ in the original optimization problem by assigning a sufficiently high cost in the cost field for points in $\mathbb{R}^2$ outside the boundaries of the ice channel $\mathcal{I}$. The following is the resulting discrete optimization problem:

\begin{mini!}|l|[2]               
    {\substack{\bm{\eta}_1,\dots,\bm{\eta}_{N+1},\\ \kappa_1,\dots,\kappa_N, \\ \Delta s}}
    {\sum_{i=1}^{N+1}\sum_{\bm{b}_j \in \mathcal{B}_d} w_j c_{\text{obs}}(\bm{g}(\bm{\eta}_i, \bm{b}_j)) \left \lVert \frac{d}{ds}\bm{g}(\bm{\eta}_i, \bm{b}_j)\right \rVert \Delta s}
    {\label{eq:optim_discrete}} 
    {}
    \breakObjective {+ L_f + \lambda \sum_{i=1}^{N}\left(\frac{\kappa_{i+1} - \kappa_i}{\Delta s}\right)^2}
    \addConstraint{\bm{\eta}_{i+1} = \bm{f}_{\text{RK4}}(\bm{\eta}_i, \kappa_i, \Delta s) \quad &   i = 1,\dots,N}
    \addConstraint{-r_{\min}^{-1} \leq \kappa_i \leq r_{\min}^{-1} \quad &   i = 1,\dots,N}  \label{eq:constraint4_d}
    \addConstraint{(x_i, y_i) \in \mathcal{I} \quad &   i = 1,\dots,N}  \label{eq:constraint5_d}
    \addConstraint{\bm{\eta}_1 = \bm{\eta}_{\textnormal{cur}}} 
    \addConstraint{\bm{\eta}_{N+1} \in \G_{\text{int}},}
\end{mini!}

This problem is solved using the NLP solver, IPOPT \cite{wachter2006implementation} (interfaced with CasADi \cite{andersson2019casadi} in our experiments), and warmed started with the initial path computed by the lattice planner stage by resampling the path to $N+1$ points. 

\section{Simulation Results}
\label{sec:sim}
\begin{figure}[t]
    \centering
    \includegraphics[width=\columnwidth]{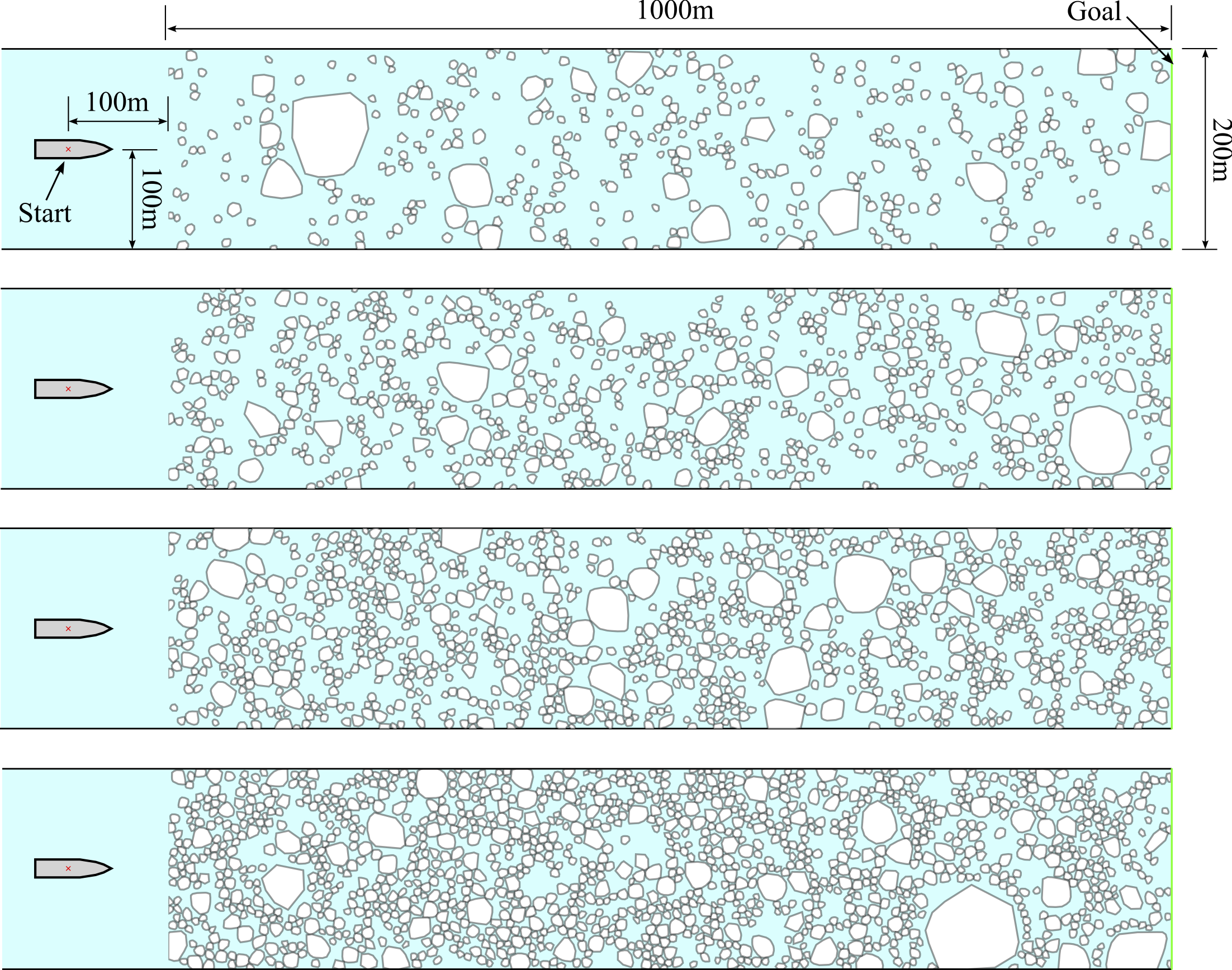}
    \caption{Sample ice fields generated for the simulation experiments. Four different ice field concentrations are shown: 20\%, 30\%, 40\%, and 50\% (top to bottom).}
    \label{fig:ice_concentrations} 
\end{figure}

Extensive simulation experiments were performed to evaluate our framework's ability to autonomously navigate a vessel in broken ice fields. The particular scenario considered is a platform supply vessel (PSV) navigating through broken ice fields of various ice concentrations and containing variable-sized medium thick first-year ice floes. We considered a narrow channel of size 1000 m $\times$ 200 m (L$\times$W) with four possible ice concentrations: 20\%, 30\%, 40\%, and 50\%. For each concentration, we generated 100 different ice fields where for each ice field we performed three trials: two baselines and the proposed method. At the start of each trial, the vessel was positioned 100 m before the beginning of the ice field, aligned with the channel length, and centered with respect to the channel width (see Fig.~\ref{fig:ice_concentrations}). A trial ended when the vessel reached the end of the ice field, a distance of 1100 m from the starting position. 
A standard PC with an Intel Core i7-7700K processor with 32 GB of memory was used for these experiments. The total wall clock time to run all simulation tests and evaluation scripts was 60 hours (not including the time to render animations). In the next sections, we describe our experimental setup in more detail, including our open-source physics simulator, followed by the simulation results. Additional details regarding the experimental setup, including the PSV model shown in Fig.~\ref{fig:ship_model} and the controller, are provided in Appendix~\ref{app:sim_details}.

\begin{figure}
    \centering
    \includegraphics[width=\columnwidth]{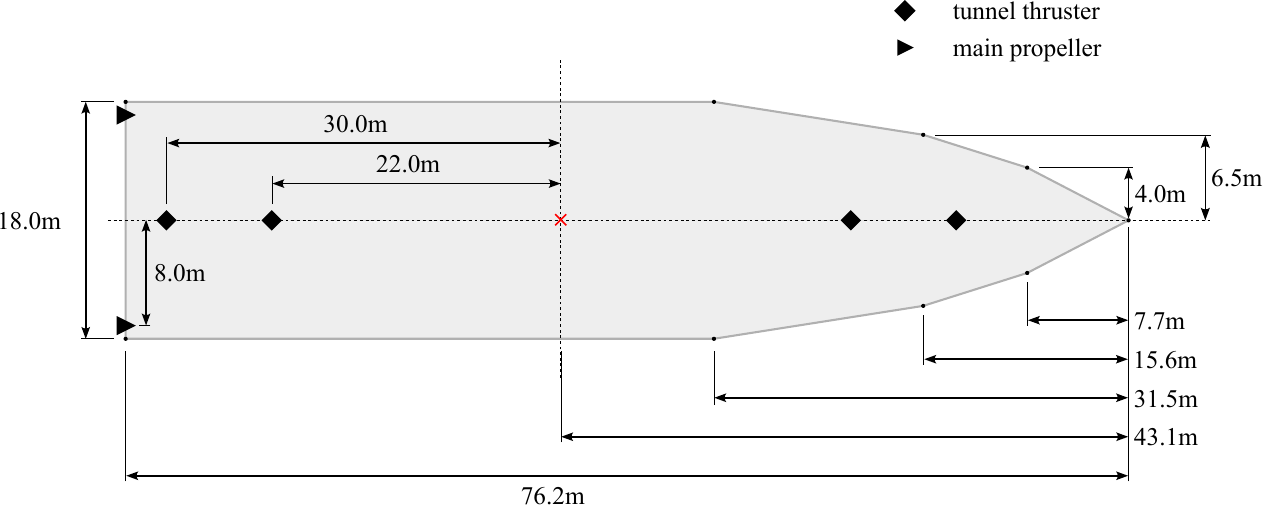}
    \caption{The dimensions, bow geometry, and actuator configuration of the vessel considered in the simulation experiments. The red `x' depicts vessel origin.}
    \label{fig:ship_model}
\end{figure}

\subsection{Physics Simulator}
Although there exist numerous works on numerical simulation of ship interactions with broken ice, e.g. \cite{huang2020ship, daley2014gpu, su2019numerical, lubbad2018overview, kim2019numerical, alawneh2015hyper, li2022review, lubbad2018simulator}, none provide open source code and few are sufficiently efficient to perform extensive experiments. We therefore developed our own simulation solution.

Our simulator is built on top of the Python library Pymunk \cite{pymunk}, which is a wrapper for the 2D real-time rigid body impulse-based physics engine Chipmunk2D \cite{chipmunk}. Collisions between bodies are therefore treated as rigid-body inelastic collisions. This means that we ignore more complex scenarios, including ice-splitting. Similar approximations have been made in other models \cite{daley2014gpu, huang2020ship}. We used existing simulation work from \cite{huang2020ship, daley2014gpu, yulmetov2016planar, su2019numerical, lubbad2018overview} and field data \cite{steer2008observed} to set the majority of the simulator parameters. Table~\ref{table:sim_parameters} in Appendix \ref{app:sim_details} summarizes these parameters and their sources.

We model the hydrodynamic force acting on the ice floes as a quadratic drag force \cite{kim2019numerical},
\begin{equation}
    \bm{F_d} =  -\frac{1}{2}\rho_w C_d A_{proj}\bm{v} \|\bm{v}\|,
\label{eq:drag}
\end{equation}
acting in the opposite direction of the ice floe linear velocity, $\bm{v} \in \mathbb{R}^2$. The water density $\rho_w$ and the drag coefficient $C_d$ are from \cite{lubbad2018overview} and \cite{su2019numerical}, respectively. The factor $A_{proj}$ is the projection of the submerged area of the ice floe along the direction of $\bm{v}$. Since our simulation is in 2D, we assume that the submerged portion of each ice floe is proportional to the density ratio $\rho_{ice} / \rho_w$. Regarding the drag torque, we model the effect on angular velocity as exponential decay.

The physics engine only handles the kinematics of the ship and therefore let the vessel dynamics account for the ice resistance forces. We do this by computing a generalized force vector $\bm{\tau}_{\text{env}} = [X_{\text{env}}, Y_{\text{env}}, N_{\text{env}}]^\top \in \mathbb{R}^3$ for a force $X_{\text{env}}$ in surge, force $Y_{\text{env}}$ in sway, and torque $N_{\text{env}}$ in yaw. This gives the net force and net torque experienced by the ship obtained by aggregating all impact forces and their corresponding contact points on the ship hull, logged during the current control interval. In practice, other environmental forces may be present in $\bm{\tau}_\text{env}$ such as wind and wave induced forces \cite{fossen2011handbook}, but we ignore these additional disturbances here. 
Although we did not validate our simulator with real-world data, we expect a reasonable level of accuracy, as the majority of our parameters and design choices are grounded in existing literature. Moreover, the average impact forces recorded in our simulations are of the same order of magnitude as those reported in prior studies \cite{daley2014gpu, lubbad2018simulator}\footnote{Some of the reported impact forces---particularly those involving large ice floes---may be inflated due to our assumptions of rigid-body collisions and the use of an impulse-based collision solver, which trades some physical realism for computational efficiency \cite{mirtich1996impulse}.}.

\subsection{Metrics}
\label{sec:sim_metrics}
The following information was stored for each simulated ship-ice collision: ice mass, collision impulse, kinetic energy loss in the system $\Delta K_{\text{sys}}$, change in kinetic energy of the ice $\Delta K_{\text{ice}}$, and contact points on the 2D ship hull. Using the logged collision data from a given trial, we computed the mean mass of the collided ice floes, the maximum magnitude of the impact force, the mean magnitude of the impact force, and the total kinetic energy loss of the ship, $\Delta K_{\text{ship}}$, using \eqref{eq:delta_Ksys2}. We approximated the impact force as the collision impulse divided by the simulation time step \cite{hasegawa2018numerical}. The impact force measurements were also used to compute the \emph{success rate} where we find the percentage of trials in which \emph{both} the mean and maximum impact forces were strictly less compared to the other navigation methods. 

The other two metrics of interest are the total energy consumption of the ship and the total transit time. Given the simulated velocities $\{\bm{\nu}_k\}_{k=1}^N$ and forces $\{\bm{\tau}_k\}_{k=1}^N$ for control steps $1,...,N$, we computed the total energy, $E$, as
\begin{equation}
    E = \Delta t_{\text{ctrl}} \sum_{k=1}^N |\bm{\nu}_k|^\top \cdot |\bm{\tau}_k|,
\label{eq:energy_use}
\end{equation}
where $|\cdot|$ denotes the element-wise absolute value operator. These additional metrics offer insight into the trade-off between efficiently traversing the ice field and altering course to avoid significant impacts with ice.

\subsection{Baselines}
We considered two baseline methods, referred to as \emph{Straight} and \emph{Skeleton}. The former plans a straight path from the initial position of the ship to the goal, while the latter refers to the shortest open water path routing approach described in \cite{gash2020machine, murrant2021dynamic}. In the first step, the Skeleton method generates a morphological skeleton image \cite{zhang1984fast} of the open water area $\mathcal{W}_{\text{free}}$. This provides a useful representation of the environment's topology from which a graph can be constructed and then searched, producing a path. The path is then post-processed with a smoothing operation, which accounts for the minimum turning radius $r_{\min}$ of the ship. We also added logic that progressively erodes the original occupancy image of the ice field until an open water path is found since no path may exist in the original image for high concentration ice fields. For a fair comparison with AUTO-IceNav, we configured the same replanning strategy\footnote{For the Straight method, the path is only planned once at the start of the trial but we still replan the velocity profile.} outlined in Algorithm \ref{alg:GeneralPlanner} for the two baselines. The parameter settings for AUTO-IceNav, including our calibration process for the parameter $\alpha$ such that \eqref{eq:nav_framework_J_cost} gives a good approximation of the true navigation cost, are given in Appendix \ref{app:param_settings_sim_experiments}. Note that the nominal speed $U_{\text{nom}}$ was set at a constant 2 m/s to facilitate the analysis across trials with different ice concentrations. 

\subsection{Results}
\label{sec:sim_results}

\begin{figure}[t]
    \centering
    \includegraphics[width=\columnwidth]{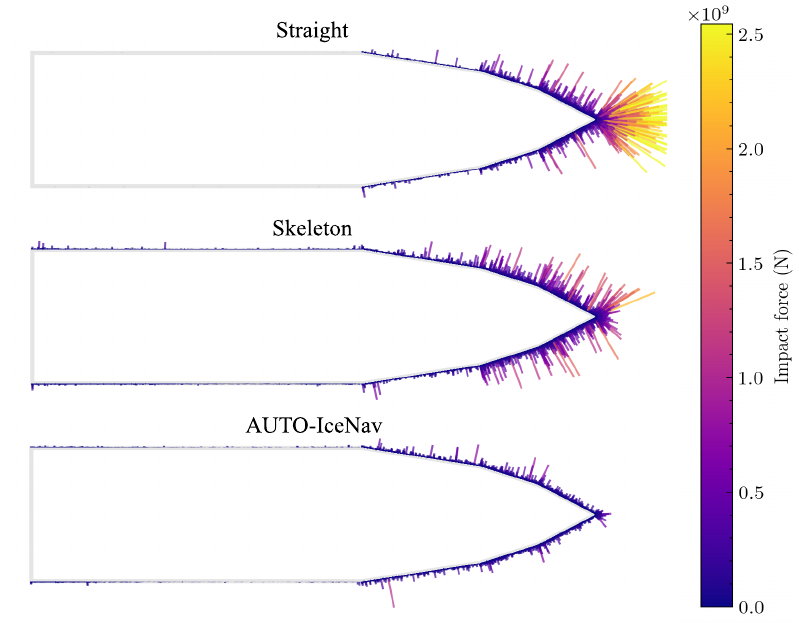}
    \caption{Visualization of all impact forces and the corresponding contact-points from the ship-ice collisions that occurred during the simulation trials. Each vector depicts the direction and relative magnitude of the impact force experienced by the ship's hull (grey outline). In total, the Straight, Skeleton, and AUTO-IceNav trials logged 11.3 million, 13.0 million, and 12.3 million collisions, respectively. 
}
    \label{fig:impacts}
\end{figure}

\begin{table*}[t]
\centering
\begin{tabular}{@{} l  l | c  c  c  c  c c c @{}}
\toprule
\multirow{3}{*}{Concentration} & \multirow{3}{*}{Method} & Mean collided & Max impact & Mean impact & Ship KE loss & Total Energy & Total time & Success \\
& & ice mass ($10^3$ kg) & force ($10^3$ kN) & force (kN) & $\Delta K_{\text{ship}}$ ($10^3$ kJ) & $E$ ($10^3$ kJ) & (s) & rate (\%) \\  
\midrule
\multirow{3}{*}{20\%} & Straight & 360 & 1025 & 576 & 100 & 245 & \textbf{604} & 0\\
 & Skeleton & 224 & 461 & 252 & 37 & 233 & 620 & 2\\
  & AUTO-IceNav & \textbf{99} & \textbf{134} & \textbf{133} & \textbf{22} & \textbf{218} & 615 & \textbf{89}\\
\midrule
\multirow{3}{*}{30\%} & Straight & 196 & 960 & 383 & 125 & 273 & \textbf{604} & 0\\
 & Skeleton & 133 & 447 & 205 & 82 & 259 & 618 & 2 \\
 & AUTO-IceNav & \textbf{82} & \textbf{147} & \textbf{140} & \textbf{41} & \textbf{236} & 614 & \textbf{89}\\  
\midrule
\multirow{3}{*}{40\%} & Straight & 168 & 936 & 394 & 147 & 316 & \textbf{613} & 0\\
 & Skeleton & 107 & 392 & 200 & 102 & 296 & 620 & 7\\
 & AUTO-IceNav  & \textbf{78} & \textbf{156} & \textbf{165} & \textbf{63} & \textbf{281} & 618 & \textbf{74}\\  
\midrule
\multirow{3}{*}{50\%} & Straight & 137 & 786 & 362 & 167 & \textbf{335} & \textbf{609} & 0\\
 & Skeleton & 99 & 360 & 235 & 128 & 363 & 638 & 8\\
 & AUTO-IceNav & \textbf{74} &  \textbf{160} & \textbf{210} & \textbf{93} & \textbf{335} & 621 & \textbf{68} \\ 
\bottomrule
\end{tabular}
\caption{Results from the simulation experiments. Scores are averaged over 100 trials for a given ice concentration and navigation method.}
\label{table:sim_results_table}
\end{table*}
A total of 1200 trials were conducted: 4 concentrations $\times$ 3 navigation strategies $\times$ 100 ice fields. The trials spanned 206 hours of simulation time logging 37 million ship-ice collisions across 1320 kilometers of ship transit. We present a summary of the results in Table \ref{table:sim_results_table}. The scores for each metric in the table are averaged over the 100 trials performed for a particular ice concentration and navigation method. In the AUTO-IceNav trials, the ship collided with smaller ice floes on average and experienced fewer head-on collisions, resulting in a significant decrease in the mean and maximum impact forces generated by the ship-ice collisions. Compared to the Straight and Skeleton trials, we reduced the mean impact force by 62\% and 27\%, respectively, and the maximum impact force by 84\% and 64\%, respectively. These improvements are best visualized in Fig.~\ref{fig:impacts}, where the force vectors are plotted at their respective contact points for all collisions recorded between the three methods. Observe the significant differences in the spatial distribution of large impacts across the ship hull and the magnitude of the force vectors between AUTO-IceNav and the two baselines. On the other hand, the success rates in the last column of Table \ref{table:sim_results_table} indicate the effectiveness of AUTO-IceNav declines as ice concentration increases, highlighting a limitation of our approach under more challenging conditions. We hypothesize that this decline in performance is due to the increased likelihood of ice floes moving in ways that cause them to intersect the planned path, rather than being pushed aside upon contact with the ship. Future work is required to further explore this trend, with more extreme ice concentrations requiring higher fidelity physics simulation to account for the pressure forces present in close pack ice (i.e. $\geq 70\%$ concentration) \cite{canada_2020}.

In addition to the reductions in impact forces, which decrease the potential for damage to the vessel, our experiments also demonstrated improvements in energy use with our proposed method. The smaller impact forces resulted in a decrease in the total kinetic energy loss of the ship from colliding with ice floes. 
Summing $\Delta K_{\text{ship}}$ across trials, AUTO-IceNav resulted in a 59\% reduction in the ship's kinetic energy loss compared to the Straight trials. This efficiency gain more than offset the additional energy required to follow the 2\% longer paths generated by AUTO-IceNav. In particular, our method achieved a decrease of 8.5\% in total energy use $E$ compared to the Straight trials. This suggests that the weighted sum of the path length and the proposed total collision cost is effective in capturing the true navigation cost of transiting a broken ice field. The guidelines established by the Canadian Coast Guard for ice navigation advise vessels to extend their journey as necessary to reduce the risk of damage which, in certain scenarios, can lead to voyages with less fuel consumption \cite{canada_2019}. In simulation, our results clearly demonstrated the feasibility of achieving both of these objectives with an autonomous navigation system. Naturally, we should also be considering the additional cost associated with longer travel time, however, this cost is likely negligible compared to the possible cost incurred from damaging the ship. Across 400 trials, totaling 67.5 hours of ship transit, AUTO-IceNav increased total travel time by just 1.5 hours compared to straight navigation, representing only a 2\% increase. It is also worth noting that the controller performed adequately in tracking the reference path at the target speed. The average cross-track error was 2.0~m, while the average heading error was 1 degree.

\subsection{Evaluating the Optimization Stage}
\begin{figure}[t]
    \centering
    \includegraphics[width=\columnwidth]{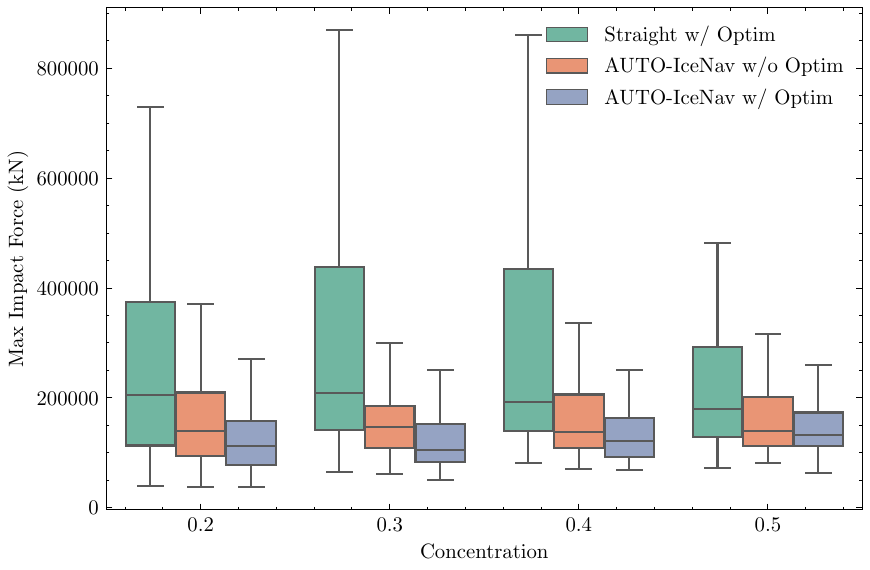}
    \caption{Results from evaluating our proposed optimization stage. Each box in the boxplot represents the distribution of maximum impact forces across 100 trials for a given ice concentration and navigation method.}
    \label{fig:lattice_optim_eval}
\end{figure}
\begin{figure}
    \centering
    \includegraphics[width=\columnwidth]{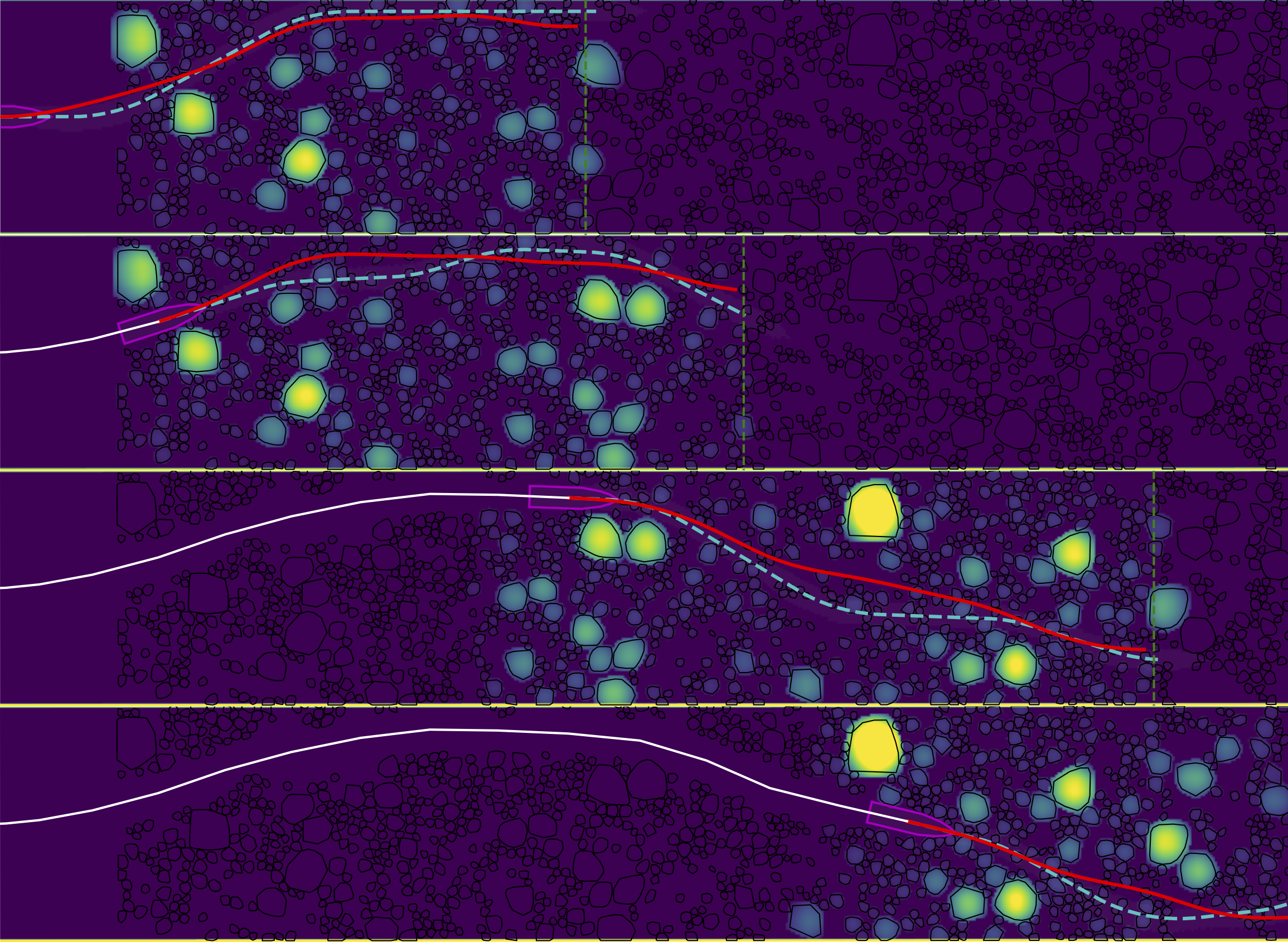}
    \caption{Four snapshots taken from a sample simulation trial in a 40\% concentration ice field using our proposed method. The initial path computed in the lattice planning stage (cyan dotted line) and the optimized path (red line) are overlaid on the costmap. Visually, we see that the red path tends to be better in terms of both the path length and the ship-ice collisions that occur along it. The actual path of the ship is displayed in white and the green dotted line shows the intermediate goal $\G_{\text{int}}$.} 
    \label{fig:lattice_optim_eval2}
\end{figure}

In addition to testing our method against the Straight and Skeleton baselines, we evaluated the effectiveness of our proposed optimization-based improvement step. We conducted the same 400 trials described above for our method \emph{without} the second stage. From these additional trials, we see that the optimization stage provides a clear improvement in navigation performance. On average, the second stage reduced the maximum and mean impact forces by 19\% and 10\%, respectively, while using 1\% less total energy compared to the AUTO-IceNav trials without optimization. As another baseline, we conducted 400 trials with the optimization stage warm-started with a straight path. These trials demonstrate that, while the optimization stage yields significant improvements, the solutions remain locally optimal. We therefore achieved better performance when the optimization stage was warm started with the lattice planning solution. Fig.~\ref{fig:lattice_optim_eval} shows the results for the maximum impact force and Fig.~\ref{fig:lattice_optim_eval2} shows four representative snapshots of a sample AUTO-IceNav trial in a 40\% concentration ice field.

\section{Results from Physical Testbed}
\label{sec:real}
\begin{figure}[t]
    \centering
    \includegraphics[width=\columnwidth]{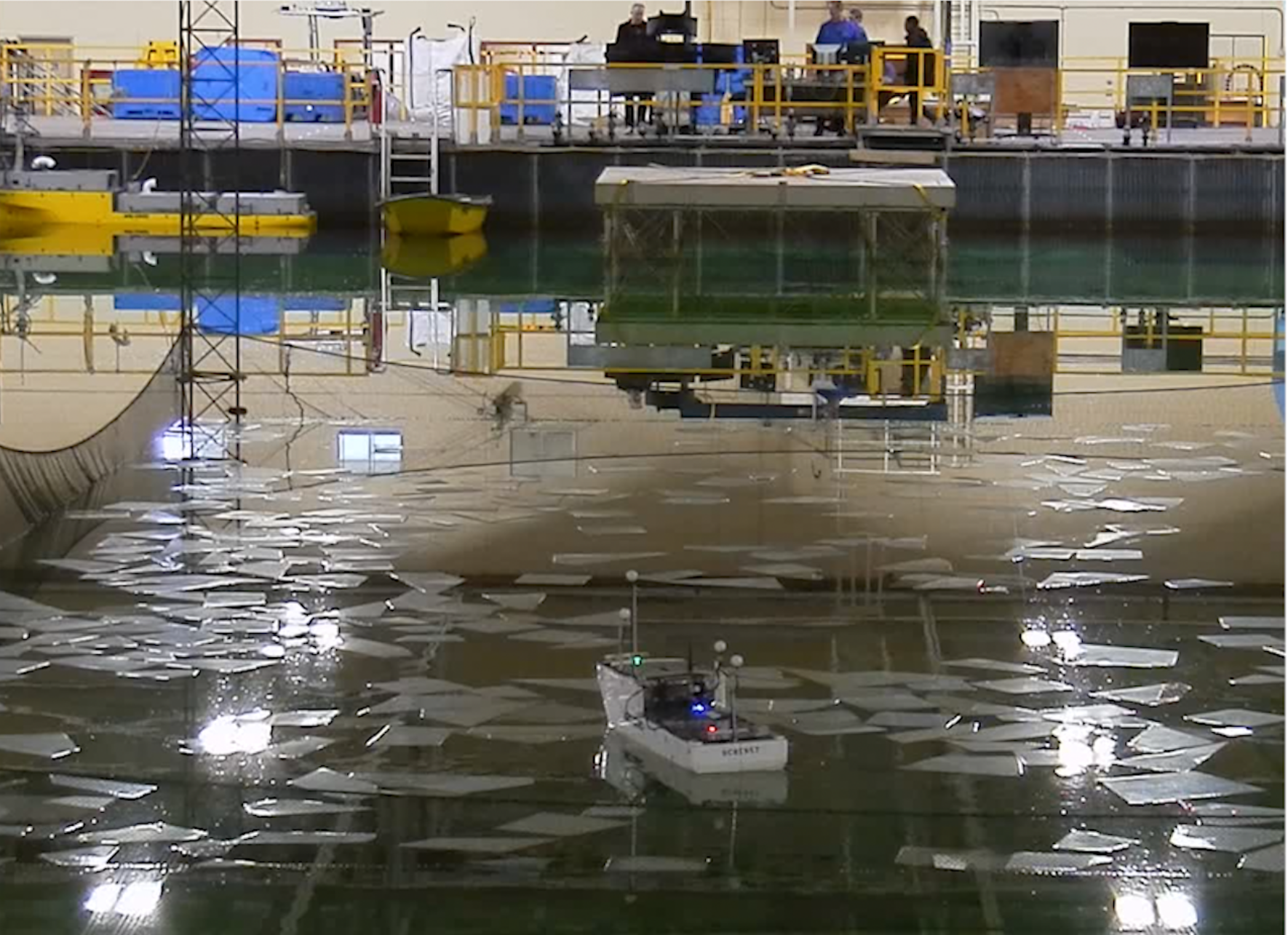}
    \caption{Side view of the experimental setup in the Offshore Engineering Basin research facility. The ice basin contains a collection of plastic ice pieces cut from sheets of polypropylene.}
    \label{fig:oeb_sample_trial}
\end{figure}

In addition to the experiments performed in simulation, we conducted real-world experiments at model-scale in the Offshore Engineering Basin (OEB) research facility in St. John's, Newfoundland and Labrador. The NRC-managed facility features a large basin measuring 75 m $\times$ 32 m, shown in Fig.~\ref{fig:oeb_sample_trial}. Our experimental setup closely followed the setup described in our preliminary work \cite{deschaetzen2023} with some necessary adjustments\footnote{The 90 m ice tank research facility used in our previous experiments was unavailable at the time of this work.}. In contrast to the NRC ice tank used in our previous experiments, the OEB is a room temperature facility. As a result, we used a collection of plastic polygon pieces cut from polypropylene sheets (density 991 kg / m$^3$ and thickness 0.012 m) commonly used for artificial ice floes \cite{ryan2021arctic}. In total, we had around 300 polygons which provided an ice concentration of 30$\%$ over a net-confined area of 20~m $\times$ 6~m (L$\times$W) for the ice channel.

The ship model used in these experiments was a 1:45 scale platform supply vessel (PSV), designed with a typical hull shape and constructed by the NRC. The dimensions of the vessel are 1.84 m $\times$ 0.38 m $\times$ 0.43 m (L$\times$W$\times$H) and weighs 90 kg. For propulsion, the PSV contains two tunnel thrusters (fore and aft) and two main propellers. More details on the PSV are described in \cite{murrant2021dynamic}. To control the vessel, we employed a Dynamic Positioning (DP) controller similar to the one used for the simulation experiments (see Appendix \ref{app:vessel_model}). The forward thrust was set constant, which achieved a nominal speed of 0.2~m/s. To capture ice information, we used a ceiling-mounted camera pointing down on the basin and used a similar image segmentation pipeline (updated at 30~Hz) described in \cite{gash2020machine}. An optical-based motion capture system (Qualisys) provided 3D (i.e. 6 DoF) state information of the vessel at 50 Hz.

\begin{figure}[t]
    \centering
    \includegraphics[width=\columnwidth]{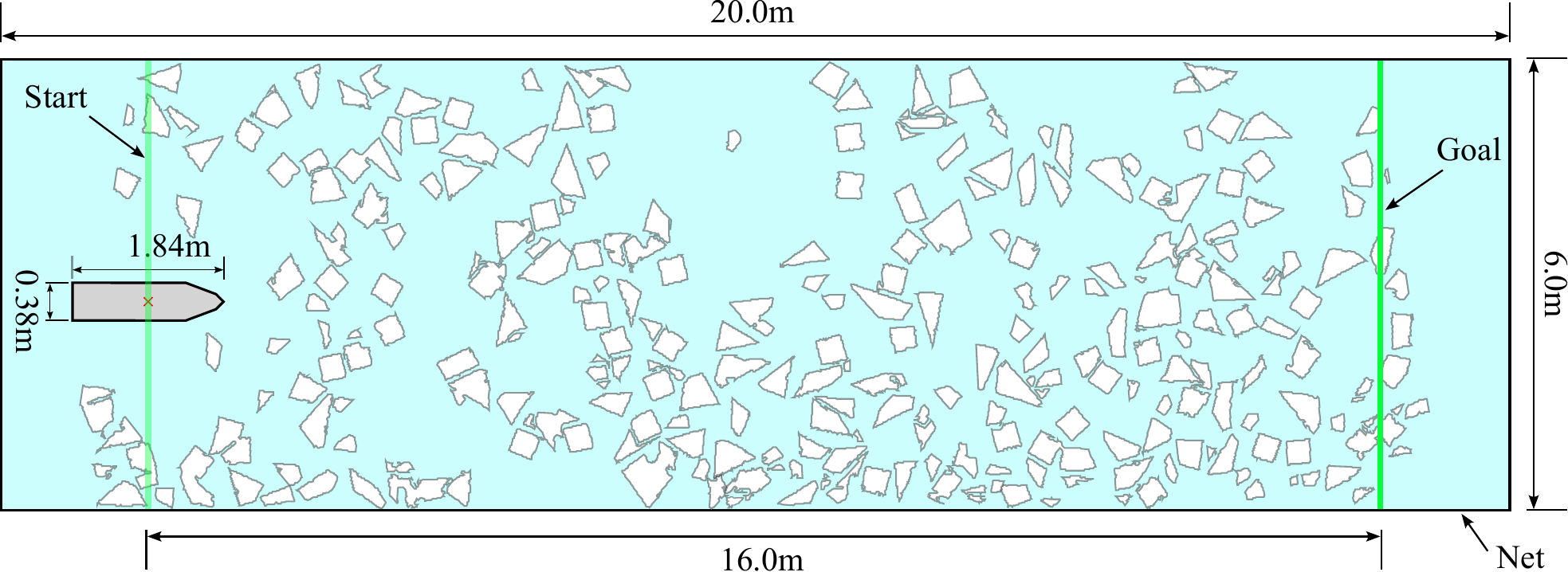}
    \caption{Diagram showing the dimensions of the model-scale ship and ice channel for our physical testbed.}
    \label{fig:oeb_setup}
\end{figure}

We performed a total of 60 trials and used the same baselines, i.e. Straight and Skeleton, from our simulation experiments to compare against our approach. Before starting each trial, the PSV was manually positioned along a starting line, with its heading aligned with the channel length. We considered three general starting positions along the starting line relative to the channel width: \emph{center}, \emph{right}, and \emph{left}. For each navigation strategy, we performed 20 trials, initializing 6 of them with a center start, 7 with a left start, and 7 with a right start. The goal line was established at the opposite end of the channel, located 16 m from the starting point. A diagram of the experimental setup is shown in Fig.~\ref{fig:oeb_setup}. Note that we used a random ordering to conduct the 60 trials.

\subsection{Metrics}
\label{sec:real_metrics}
The small-scale setup limited our ability to rely on several of the metrics discussed in the simulation section, including impact forces derived from acceleration data measured by the onboard IMU. We computed an approximation for $\Delta K_{\text{ship}}$ by performing object tracking on the ice segmentation data. A segmented ice floe is represented as a polygon, defined by a sequence of vertices at time $t_k$, where $k \in {1, \dots, N}$ denotes the time steps of the trial. Each polygon was tracked using a Kalman filter to estimate the position $\bm{p} \in \mathbb{R}^2$ and the velocity $\bm{v} \in \mathbb{R}^2$ of the polygon's centroid. Using the velocity estimates, we identified the subset of segmented ice floes, $\mathcal{Q}$, that were likely pushed by the ship during a particular trial. This included scenarios where an ice floe was in direct contact with the ship or was touching a chain of other ice floes that were all being pushed by the ship. From here, we computed several approximations for the \emph{work} done on the ice by the ship, which is an equivalent definition of $\Delta K_{\text{ice}}$ when the ice floes are rigid bodies. This provides a reasonable way to compute $\Delta K_{\text{ship}}$ using \eqref{eq:delta_Ksys2} where we treat $\Delta K_{\text{sys}}$ as negligible.

The work $W$ of a net force $\bm{F}$ applied to a particle with mass $m$ is given by
\begin{equation}
    W = \int_{t_1}^{t_2} \bm{F} \cdot \bm{v}dt = m \int_{t_1}^{t_2} \bm{a} \cdot \bm{v} dt,
\end{equation}
where $\bm{a}$, $\bm{v}$ is the particle's acceleration and velocity, respectively. We considered the net force acting on the ice floes to consist of the drag force and the force from the ship (either via direct contact or from a chain of ice floes).
This means that in the absence of a ship force, work done on the ice will be negative if the ice floe speed $\|\bm{v}\|$ is nonzero. We can thus approximate the work done by the ship on the ice by only considering the instances of positive work in the object tracking data. We used the velocity estimates $\{\bm{v}^P_k\}_{k=1}^N$ for each pushed ice floe $P \in \mathcal{Q}$ tracked by the Kalman filter to give us the total positive work,
\begin{equation}
    W_1 = \sum_{P \in \mathcal{Q}} m^P \sum_{k=1}^{N-1} (t_{k+1}-t_{k})\max(0, \bm{a}^P_{k} \cdot \bm{v}^P_{k}).  
\label{eq:work1}
\end{equation}
Note, the ice mass $m^P$ is computed as the product of ice thickness, ice density, and area. Acceleration $\bm{a}_k^P\in \mathbb{R}^2$ was approximated via finite differencing of the velocity estimates. If we ignore the work associated from a resultant torque, then \eqref{eq:work1} gives a good approximation for the positive work done by the resultant force applied on a rigid body.

We wanted to see whether a coarser metric (i.e., worse approximation but perhaps less prone to noise) for work would produce similar results as \eqref{eq:work1}. Hence, instead of considering an accumulation of positive work, we considered the change in kinetic energy for each polygon $P \in \mathcal{Q}$ given the maximum speed attained $v^P_{\max} = \max{\{\|{\bm{v}}_k^P\|\}_{k=1}^N}$ and the initial speed $\|\bm{v}_1^P\|$. Using the work-energy principle, we get our second metric $W_2$ for the work done by the ship on the ice:
\begin{equation}
    W_2 = \frac{1}{2} \sum_{P \in \mathcal{Q}} m^P ((v^P_{\max})^2 - \|\bm{v}_1^P\|^2).
\label{eq:work2}
\end{equation}
Observe that \eqref{eq:work2} does not capture the scenario where an ice floe was pushed on several distinct occasions by the ship meaning $W_1 \geq W_2$. As a second coarse metric for work, we computed the product of the ice mass $m^P$ and the arc length $s^P$ of the path $\{\bm{p}^P_k\}_{k=1}^N$ taken by the polygon $P\in\mathcal{Q}$:
\begin{equation}
    W_3 = \sum_{P \in \mathcal{Q}} m^P s^P.
\label{eq:work3}
\end{equation}
While \eqref{eq:work3} lacks a direct connection to work, it offers a straightforward way for capturing the amount by which the ship changed the ice environment.

Other metrics we computed from the object tracking data are the number of pushed ice floes given by the cardinality of the set $\mathcal{Q}$ and the mean mass of the set of ice floes that collided with the ship. The latter metric was computed by finding the subset of polygons $\mathcal{Q'} \subseteq \mathcal{Q}$ that intersected the ship footprint at some point during the trial. We also included the total time to transit the ice channel.

\subsection{Results}
\begin{table*}[t]
\centering
\begin{tabular}{@{} l  l | c  c  c  c  c  c  c@{}}
\toprule
Difficulty & Method & Mean collided ice mass (kg) & No. pushed ice & $W_1$ (J)  & $W_2$ (J) & $W_3$ &  Total time (s) \\
 % work 1 is path length x mass
 % work 2 is sum delta KE
 % work 3 is sum integral net force v dt
\midrule
\multirow{3}{*}{Easy} 
 & Straight & 1.28 & 44.13 & 0.28 & 0.20 & 40.91 &  \textbf{91.39}\\
 & Skeleton & 1.21 & 128.00 & 0.75 & 0.47 & 127.11 & 115.27\\
 % & Lattice  & 15.67 & \textbf{0.65} & 1.36 & 10.65 & 49.33 & \textbf{0.26} & \textbf{0.18} & 46.92\\  % lattice2
  & AUTO-IceNav & \textbf{1.15} & \textbf{35.00} & \textbf{0.20} & \textbf{0.15} & \textbf{28.87} &  101.91\\  % lattice1
\midrule
\multirow{3}{*}{Medium} 
 & Straight & 1.27 & 68.40 & 0.44 & 0.31 & 65.72 &  \textbf{89.52} \\
 & Skeleton & 1.31 & 102.63 & 0.59 & 0.38 & 106.55 & 110.06 \\
 % & Lattice  & \textbf{16.50} & 0.70 & 1.55 & \textbf{11.53} & \textbf{45.93} & \textbf{0.22} & \textbf{0.15} & \textbf{36.73} \\
 & AUTO-IceNav & \textbf{1.17} & \textbf{38.00} & \textbf{0.21} & \textbf{0.13} & \textbf{28.60} &  102.90 \\  % lattice1
\midrule
\multirow{3}{*}{Hard} 
 & Straight & 1.22 & 109.14 & 0.85 & 0.52 & 144.69  & \textbf{97.15}\\
 & Skeleton & 1.24 & 97.11 & 0.64 & 0.41 & 109.82 &  110.73\\
 % & Lattice  & \textbf{20.33} & 0.66 & \textbf{1.27} & \textbf{13.24} & \textbf{47.00} & \textbf{0.25} & \textbf{0.16} & \textbf{41.48}\\
 & AUTO-IceNav  & \textbf{1.20} & \textbf{49.33} & \textbf{0.33} & \textbf{0.21} & \textbf{53.86} &  109.35\\  % lattice1
\bottomrule
\end{tabular}
\caption{Results from experiments done in the physical testbed.}
\label{table:metrics_table}
\end{table*}

\begin{figure}[t]
  \centering
  \includegraphics[width=\columnwidth]{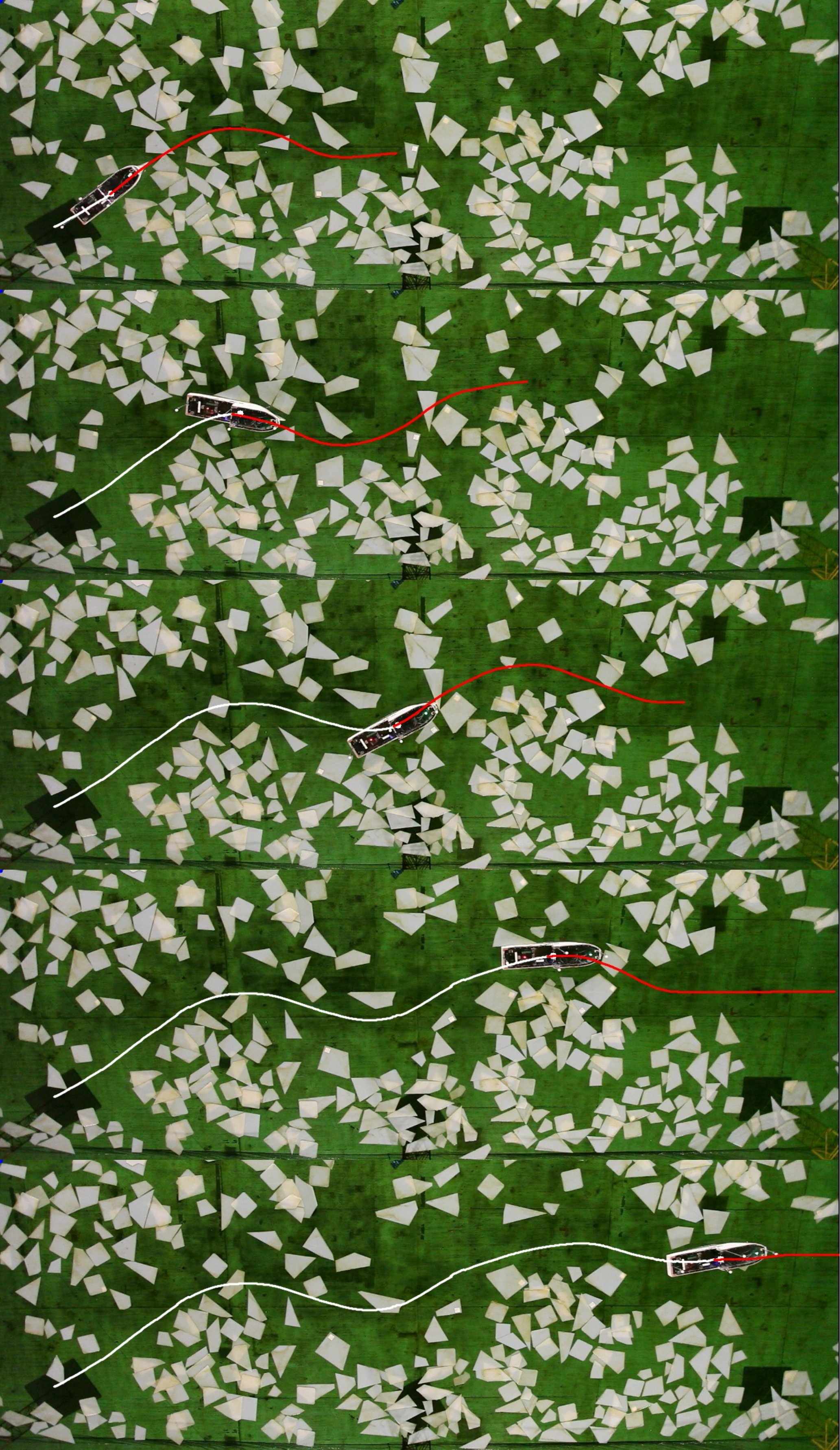}
  \caption{Five snapshots from a sample AUTO-IceNav trial, displaying the planned path in red and the actual ship path in white. Videos of trials can be found at {\color{magenta}\url{https://github.com/rdesc/AUTO-IceNav\#demo-videos-from-nrc-experiments}}.}
  \label{fig:oeb_snapshots}
\end{figure}

We present a summary of the results in Table \ref{table:metrics_table} and show several snapshots of a sample AUTO-IceNav trial in Fig.~\ref{fig:oeb_snapshots}. The table shows the metrics discussed in Section \ref{sec:real_metrics}, where the scores are averaged across trials for each navigation method and difficulty level. The three levels, `easy', `medium', and `hard', are based on the average subjective difficulty scores assigned to each trial by the authors as part of a blind evaluation process (see Appendix~\ref{app:trial_difficulty} for more details).

\begin{figure}[!t]
    \centering
    \includegraphics[width=\columnwidth]{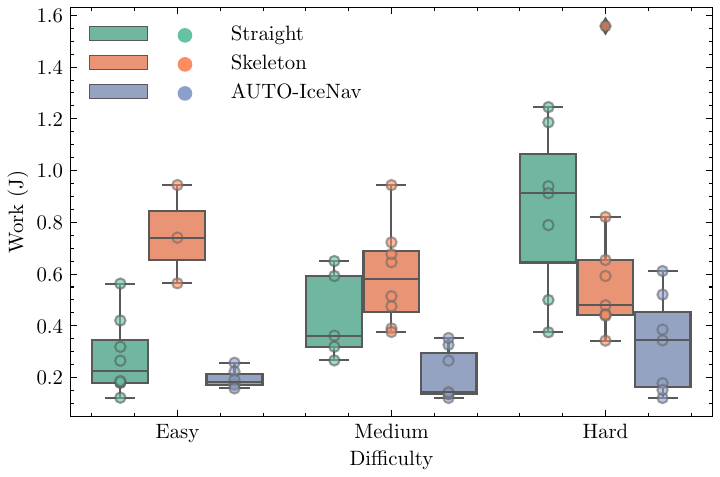}
    \caption{Total work done on the ice compared across method and difficulty level. Work was computed using the object tracking data and \eqref{eq:work1}.}
    \label{fig:oeb_hist_net_force}
\end{figure}

In terms of work, our approach outperformed the baselines in the three difficulty levels. Compared to the second-best score for $W_1$ from \eqref{eq:work1}, we achieved improvements of $29\%$, $52\%$, and $48\%$ for the easy, medium, and hard tests, respectively. These relative improvements are similar across the other two work metrics, suggesting that even the coarser metric $W_3$ provides a useful measure of performance. Fig.~\ref{fig:oeb_hist_net_force} provides a complete picture of the results for $W_1$. Observe that for the easy trials, our method provided the smallest gains over navigating straight through the ice field. This gap between Straight and AUTO-IceNav grows as the trials become more difficult. In addition, Straight is outperformed by both AUTO-IceNav and Skeleton for the hard trials. Interestingly, the Skeleton approach exhibits similar performance across difficulty levels and performed worse on average for the easy trials. However, it is difficult to draw definitive conclusions due to the limited number (only 3) of easy Skeleton trials. The results also indicate that AUTO-IceNav enabled the ship to, on average, collide with smaller ice floes and interact with fewer polygons.

We highlight several limitations of these experiments. Due to resource constraints, we were unable to perform trials with varying ice concentrations like we did in the simulation experiments. It would be interesting to see how our improvements over the baselines compare for higher concentration ice fields, e.g. 40\% and 50\% ice concentration. We hypothesize that such experiments would show similar performance trends compared to the simulation results described in Section~\ref{sec:sim_results}. However, in contrast to the simulation experiments, the plastic ice floes had a narrower range in size and shape, which may have limited the range of possible scenarios encountered during ship transit. At a broader level, the small-scale experiment setup and relatively lightweight fake ice made it challenging to simulate full-scale behavior. For instance, the acceleration measurements from the onboard IMU were not particularly meaningful, as mentioned earlier. We were also unable to make meaningful comparisons in total energy consumption due to the lack of resistance provided by the fake ice. This means that the computed work scores seem less significant if we compare them with the total energy used by the ship actuators. Future experiments in the physical testbed would, therefore, benefit from an increase in the ice concentration and the ice floe thickness. 


\section{Conclusion}

In this work, we examined the problem of local navigation for autonomous ships operating in broken ice fields. We presented our framework, AUTO-IceNav, which we showed to be an effective navigation strategy both in simulation and in a physical testbed. Our results demonstrated significant reductions in the mean and maximum impact forces from ship-ice collisions and decreased energy consumption compared to naive navigation strategies such as navigating straight.

\subsection{Limitations and Future Work}

\subsubsection{Oscillating between similar cost paths}
The large number of obstacles in the environment, along with the exponential number of possible ship-ice interaction combinations, can give rise to multiple---often many---locally optimal paths with similar costs. In practice, this manifests as oscillations between multiple paths of similar cost across planning iterations. We mitigated this effect by applying a threshold on the cost difference in selecting between the new and old paths. A better approach would be to find the various modes in parallel, clustering paths using a notion of homotopy class and using a more robust strategy for ensuring consistency across planning iterations.

\subsubsection{Handling noisy ice measurements}
During the physical experiments, we observed that the ice segmentation pipeline is prone to producing noisy outputs, affecting downstream processing. This suggests the use of a probabilistic representation of the ice field where measurements are used to iteratively update beliefs on the shape, position, and orientation of the ice floes. Such a probabilistic approach would be particularly useful in accounting for uncertainties inherent in ship-mounted sensors~\cite{zhang2022semantic}. This would also allow us to incorporate the notion of \emph{risk} into our objective, where we can use the estimates and the associated uncertainty of the ice data to penalize high-risk scenarios.

\subsubsection{More accurate collision cost}
Higher-fidelity models for ship-ice collisions are described in \cite{daley1999energy, daley2012gpu, daley2014gpu}, which account for additional important features that affect the collision physics, including the angles of the ship's hull, the shape of the ice edges, and the pressure effects in close pack ice conditions. However, such models do not translate well with our current costmap-based approach and, therefore, would require novel solutions for designing a suitable cost function for efficient planning. Even more challenging would be to account for the cost from breaking through the ice, which typically occurs during head-on collisions with larger ice floes \cite{li2022review}. Another improvement would be to add constraints on the maximum impact force experienced during ship-ice collisions, as it has a greater effect on structural safety than overall energy \cite{li2022review}.

\subsubsection{Handling the changing ice field}
A major direction for future work is to account for the changing ice field during the planning stage. 
This means that instead of relying on frequent replanning, we use a prediction model of the ice field to better inform planning behavior, particularly in higher-concentration ice conditions, where ship-induced ice motion can lead to suboptimal interactions that degrade navigation performance without accurate modeling of ice field dynamics. This suggests that a useful ice prediction model should receive as input information that encodes how the ship will navigate the current environment. Modeling the relationship between the state of the ice field, the action of the ship, and the navigation cost may be a suitable problem for reinforcement learning algorithms and other learning-based methods \cite{zhong2025autonomous}.

% Finally, another major improvement to our method would be to treat the ship's velocity as variable rather than keeping it constant when calculating the total travel time and total collision cost. Such an approach would likely result in substantial performance gains in terms of navigation metrics and could enable a wider range of desirable ship maneuvers for ice navigation, including slowing down before large collisions, as necessary.


%%%%%%%%%%%%%%%%%%%%%%%%%%%%%%%%%%%%%%%%%%%%%%%%%%%%%%%%%%%%%%%%%%%%%%%%%%%%%%%%

%%%%%%%%%%%%%%%%%%%%%%%%%%%%%%%%%%%%%%%%%%%%%%%%%%%%%%%%%%%%%%%%%%%%%%%%%%%%%%%%

%%%%%%%%%%%%%%%%%%%%%%%%%%%%%%%%%%%%%%%%%%%%%%%%%%%%%%%%%%%%%%%%%%%%%%%%%%%%%%%%
\appendices
\section{Search Heuristic}
\label{app:appendix_A_heuristic}

Given the two terms in \eqref{eq:nav_framework_J_cost}, we compute two separate heuristics---each admissible with respect to its corresponding term in the cost function---to improve the runtime of A* search during lattice path planning. Specifically, we require a function $h_1(\bm{\eta})$ such that it lower bounds the path length $L_f$ and another function $h_2(\bm{\eta})$ such that it lower bounds the total collision cost $C_f$ for all $\bm{\eta}$ given the current goal $\G_{\text{int}}$ and costmap $\mathcal{C}_{\textup{map}}$. We can therefore define an admissible heuristic function $h(\bm{\eta})$ as the weighted sum:
\begin{equation}
h(\bm{\eta}) = h_1(\bm{\eta}) + \alpha h_2(\bm{\eta}).
\end{equation}
We refer to the two components of $h$ as the \emph{nonholonomic without obstacles} heuristic and the \emph{obstacles-only} heuristic, respectively.

\subsection{Nonholonomic without obstacles}
Path length $L_f$ is lower bounded by the length of the shortest path from $\bm{\eta}$ to an infinite line $\G_{\infty}$ that is collinear with the intermediate goal $\G_{\text{int}}$, subject to the unicycle model \eqref{eq:unicycle}. Recall our definition of the goal in \eqref{eq:goal} and recall our world coordinate frame shown in Fig.~\ref{fig:prob_formulation}. To characterize the proposed heuristic, we offer the following result.

\begin{figure}[!t]
    \centering
    \includegraphics[width=\columnwidth]{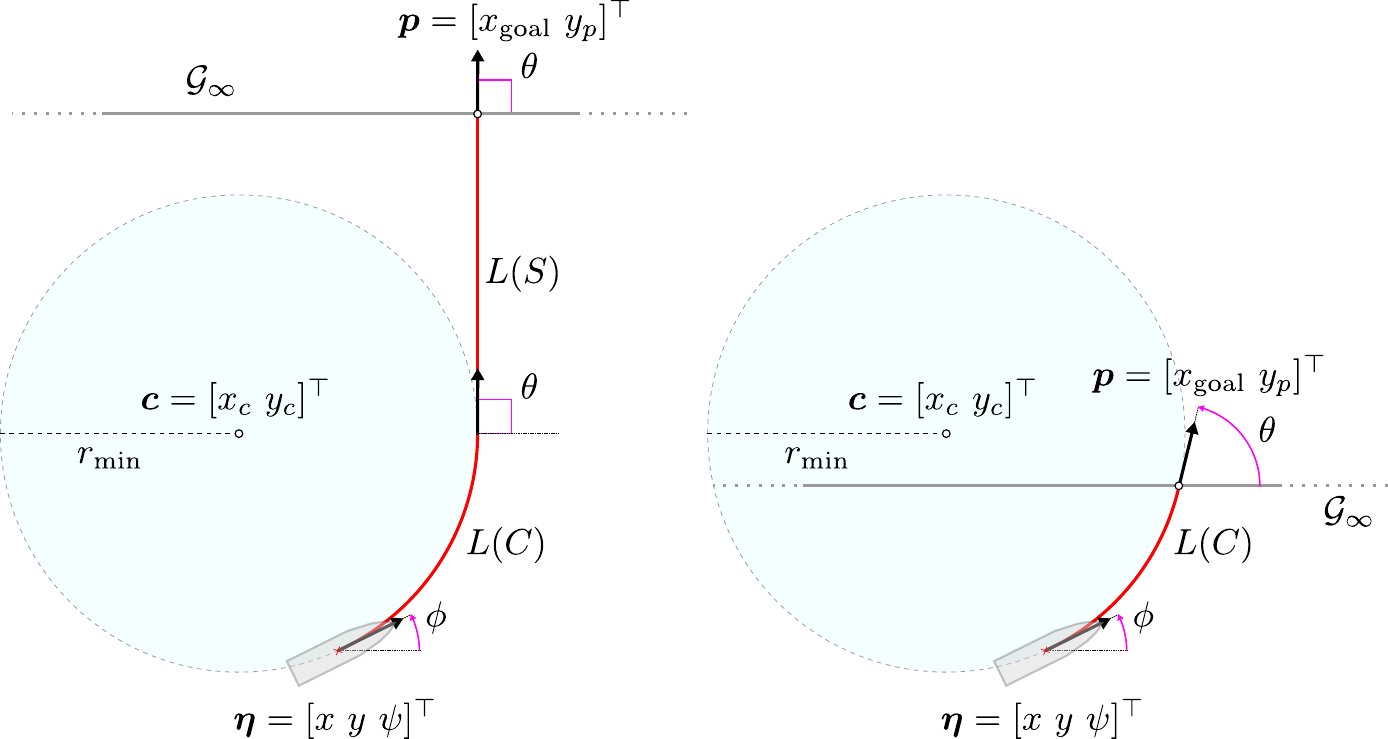}
    \caption{Diagram depicting the geometry of the Dubins path to an infinite line $\G_{\infty}$ for two possible cases.}
    \label{fig:heuristic}
\end{figure}

\begin{theorem}[Closed-form Heuristic for Path Length]
\label{theorem:1}
The shortest path from $\bm{\eta} = [x \ y \ \psi]^\top$ to the infinite straight line segment $\G_{\infty}$ with minimum turning radius $r_{\min}$ is a Dubin's path. Referencing Fig.~\ref{fig:heuristic}, 
\begin{enumerate}[label=(\alph*)]
    \item  if $\G_{\infty}$ lies above the point $\bm{c} = [x_c \ y_c]^\top$, the path is of the form CS (circular arc C of radius $r_{\min}$, followed by a straight line segment S) and S intersects $\G_{\infty}$ at $\theta = \pi / 2$ (Fig.~\ref{fig:heuristic} (left));
    \item otherwise, the path is of the form C (Fig.~\ref{fig:heuristic} (right)).
\end{enumerate}
Given these two cases, $h_1(\bm{\eta})$ is the path length and is given analytically by:
\begin{equation*}
    \begin{split}
    h_1(\bm{\eta}) &=\begin{cases}
    h_a(\bm{\eta}), \ &\text{if } x_c \leq x_{\text{goal}} \ \text{(case a)},\\
    h_b(\bm{\eta}), \ &\text{otherwise (case b)}
    \end{cases},\\
    \text{where} &\\
         h_a(\bm{\eta}) &= r_{\min}\left|\frac{\pi}{2} - \phi\right| + x_{\text{goal}} - x_c\\
         h_b(\bm{\eta}) &= r_{\min}\left|\phi - \arccos\left(\frac{x_c - x_{\text{goal}}}{r_{\min}}\right)\right|\\
         x_c &= x + r_{\min} \left|\cos(\phi)\right|\\
         % y_c &= y + r_{\min} |\sin(\phi)|\\
    \end{split}
\end{equation*}\\
and where $\phi \in [-\frac{\pi}{2}, \frac{\pi}{2}]$ is the heading measured with respect to the line segment $\G_{\infty}$.
\end{theorem}

\begin{proof}

Consider a circle of minimum turning radius $r_{\min}$ that is tangent to $\bm{\eta}$. The position $\bm{c} = [x_c \ y_c]^\top$ is denoted as the center of this circle.
Without loss of generality, we consider a left turn at an angle of $\phi \in [0, \frac{\pi}{2}]$ as in Fig.~\ref{fig:heuristic} (a similar analysis can be made for right turns and $\phi \in [-\frac{\pi}{2}, 0]$ --- heading angles that point `down' with respect to $\G_{\infty}$).

\emph{Case 1:}
Suppose that $\G_\infty$ lies above $\bm{c} = [x_c \ y_c]^\top$. This is equivalent to the condition that $x_c$ is less than $x_{\text{goal}}$, i.e.
\begin{equation}
    x_c = x + r_{\min}\cos(\phi)\leq x_{\text{goal}}.
\end{equation} Let $\bm{p} = [x_{\text{goal}} \ y_p]^\top$ be any point on $\G_{\infty}$ and is outside of the circle of radius $r_{\min}$ centered at $\bm{c}$ (it is trivial to show the point $\bm{p}$ cannot be inside this circle for the shortest path from $\bm{\eta}$ to $\G_{\infty}$). Since the heading of $\bm{p}$ is not specified, the shortest path from $\bm{\eta} = [x \ y \ \psi]^\top$ to $\bm{p} = [x_{\text{goal}} \ y_p]^\top$ is of the form CS~\cite{bui1994shortest,ma2006receding} where the straight line segment $S$ intersects $\G_{\infty}$ at an angle $\theta$. Therefore, determining the shortest path from $\bm{\eta}$ to $\G_{\infty}$ reduces to computing
\begin{equation}
    \min_{\theta} \quad L(C)+L(S),
\end{equation} where $L(\cdot)$ denotes length. We observe 
\begin{equation}
\begin{split}
L(C) &= r_{\min}(\phi - \theta),\\
L(S) &=  \frac{x_{\text{goal}} - [x + r_{\min}(\cos(\phi) - \cos(\theta))]}{\sin(\theta)}.
\end{split}
\end{equation}
Further, we observe that the total path length $L(C)+L(S)$ is minimized by $\theta=\pi/2$. Replacing this value in $h_a(\bm{\eta})=L(C)+L(S)$ yields the result of the Theorem for the case $x_c \leq x_{\text{goal}}$.

\emph{Case 2:} Suppose instead that $\G_\infty$ lies below $\bm{c}$. In this case, the shortest path from $\bm{\eta}$ to $\G_\infty$ consists only of a circular arc $C$ which intersects $\G_\infty$ and has length equal to the result of the Theorem for the case that $x_c > x_{\text{goal}}$.
\end{proof}

\subsection{Obstacles-only}
The following is an admissible heuristic $h_2(\bm{\eta})$ for $\eqref{eq:swath_cost}$, i.e., a lower bound for the collision cost of the swath $\mathcal{S}(\bm{\pi})$ given a pose $\bm{\eta}$ and the current goal $\G_{\text{int}}$. Let $w \in \mathbb{N}$ be the width of the ship footprint in costmap grid units. Now suppose that we have a path $\bm{\pi}$, starting from $\bm{\pi}(0) = \bm{\eta}$ and ending at $\bm{\pi}(L_f) \in \G_{\text{int}}$, and a corresponding swath $\mathcal{S}(\bm{\pi})$ that sweeps rows $i$ through $j$ in the costmap for a total of $N = j - i + 1$ rows. It must hold that the part of the swath that covers a particular row must consist of a set of grid cells $\mathcal{S}' \subset \mathcal{S}(\bm{\pi})$ with at least $w$ consecutive elements, for all $N$ rows. In other words, we have $|\mathcal{S}'| \geq w$ in the general case and $|\mathcal{S}'| = w$ when the ship heading $\psi$ is aligned with the length of the channel / costmap. A lower bound for the cost of the single-row swath $\mathcal{S}'$ for any particular row $k$ in the costmap is therefore achieved by selecting the lowest sum of $w$ consecutive grid cells in the $k$-th row. If we extend this to all $N$ rows in the costmap that make up the swath $\mathcal{S}(\bm{\pi})$, then we get a lower bound for the cost of the swath $\mathcal{S}(\bm{\pi})$ for all paths $\bm{\pi}$ from $\bm{\eta}$ to $\G_{\text{int}}$ and for all $\bm{\eta}$. Although this heuristic relies on a loose lower bound on the cost, it is efficient to compute (linear search for each row) and admissibility is guaranteed.

\section{Comparison with Body Points in Zucker et al.}
\label{app:body_points}
\begin{figure}[t]
    \centering
    \includegraphics[width=\columnwidth]{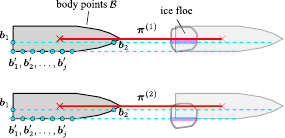}
    \caption{A simple scenario to illustrate the problem with defining body points exclusively on the outline of the ship body. Two candidate paths are considered, $\bm{\pi}^{(1)}$ and $\bm{\pi}^{(2)}$, in the top and bottom parts of the figure, respectively. We examine the cost of these two paths by considering the contribution to the cost from two subsets of $\mathcal{B}$: $\{\bm{b}_1, \bm{b}_2\}$ and $\{\bm{b}_1', \bm{b}_2', \dots, \bm{b}_j'\}$, depicted as cyan dots. The corresponding paths of the body points are shown with the cyan dotted lines. We see that in the bottom scenario, the magenta region of the cost field is disproportionately penalized from $\{\bm{b}_1', \bm{b}_2', \dots, \bm{b}_j'\}$ compared to the top scenario.}
    \label{fig:body_points_scenario}
\end{figure}
The total collision cost $C_f$ in $\eqref{eq:optim_step_total_collision_cost}$ shares similarities with the obstacle objective in the collision-free trajectory optimization approach from \cite{zucker2013chomp}. However, in addition to optimizing a collision-prone path rather than a collision-free trajectory, our cost function $c_{\text{obs}}$ does not assign costs in the obstacle-free region $\mathcal{W}_{\text{free}}$ given $\obs$ in contrast to the obstacle objective in \cite{zucker2013chomp}. This means that we need to define the body points slightly differently.

To see why this is the case, consider the simple scenario illustrated in Fig.~\ref{fig:body_points_scenario}. It consists of an ice floe $\mathcal{O}$ and two candidate straight paths $\bm{\pi}^{(1)}$, $\bm{\pi}^{(2)}$, both of which result in a collision with $\mathcal{O}$ where $\bm{\pi}^{(1)}$ results in a more head-on collision than $\bm{\pi}^{(2)}$. Now suppose that the set of body points $\mathcal{B}$ is defined along the exterior of the body as in \cite{zucker2013chomp}. In particular, consider the following two subsets of $\mathcal{B}$: $\{\bm{b}_1$, $\bm{b}_2$\} located at the stern and bow, respectively, and $\{\bm{b}_1'$, $\bm{b}_2', \dots, \bm{b}_j'\}$ positioned along one of the straight sides of the ship body. We want to compare the total collision cost $C_f$ from \eqref{eq:optim_step_total_collision_cost} for the two paths by examining how $\mathcal{B}$ affects this cost. Observe that for path $\bm{\pi}^{(2)}$ each body point in $\{\bm{b}_1'$, $\bm{b}_2', \dots, \bm{b}_j'\}$ accumulates a cost from the region in the cost field highlighted in magenta. By comparison, consider how this cost field region affects the cost of $\bm{\pi}^{(1)}$. Here, there are exactly two body points, $\bm{b}_1$ and $\bm{b}_2$, which accumulate cost in this region. This suggests that the cost computed from \eqref{eq:optim_step_total_collision_cost} will be higher for $\bm{\pi}^{(2)}$ compared to $\bm{\pi}^{(1)}$. However, if we recall the original swath cost formulation for $C_f$ in \eqref{eq:swath_cost}, we know that $\bm{\pi}^{(1)}$ should be assigned a higher cost between these two candidate paths.

As a result, as shown in Fig.~\ref{fig:body_points_scenario}, defining body points exclusively on the outline of the ship body leads to cases where certain subsets of $\mathcal{B}$ disproportionately accumulate cost from our cost field. Such a $\mathcal{B}$ would create an incentive to reduce the cost of a path by slightly adjusting it such that the interior region of the ship body now passes through a high-cost region that was part of the original path.

\section{Physics Simulator Details}
\label{app:sim_details}

\subsection{Ice Field Generation}
\label{app:ice_field_gen}

Our ice field generation method produces random ice fields given a specified ice concentration where the ice floe statistics are consistent with the literature.  Following \cite{daley2014gpu, huang2022new}, we modelled the ice mass as a random variable that follows a log-normal distribution with probability density function $f_Y(y)$ given by
\begin{equation}
    \begin{split}
       f_Y(y) &= \frac{1}{b}f_X\left ( \frac{y - a}{b} \right) \\
       f_X(x) &= \frac{1}{x \sigma \sqrt{2 \pi}} \exp \left [ - \frac{(\ln x)^2}{2 \sigma^2}\right].
    \end{split}
\end{equation}
The location-scale parameters, with values $a = 10.21$ and $b = 0.9324$, are from \cite{daley2014gpu} and we set $\sigma = 0.54$ (not provided in \cite{daley2014gpu}) to achieve a similar mean and standard deviation as in \cite{daley2014gpu, huang2022new}. Using the set of sampled mass values $\{m_i\}_{i=1}^N$ (in kg), we calculated the corresponding floe area $A_i$ given the ice density $\rho_{ice}$ and the thickness $h$. We restricted our attention to ice floes with \emph{effective width} (i.e. the square root of area) between 4~m and 100 m, a reasonable range for small ice floes \cite{steer2008observed}. To generate a random arrangement of ice floes, we first applied the circle packing algorithm described in \cite{wang2006visualization} with the set of radii calculated from the set of floe sizes $\{A_i\}_{i=1}^N$ as input. The arrangement of nonoverlapping circles was generated such that it covered the whole ice field. From here, the centers and radii of the circles were used to initialize the algorithm from \cite{valtr1995probability} to generate random convex polygons consisting of 5 to 20 sides \cite{daley2012gpu}. In the final step, we randomly removed polygons from the ice field until we reached the target ice concentration. 

We generated 100 ice fields for each of the four ice concentrations considered in our experiments and show an example for each in Fig.~\ref{fig:ice_concentrations}. Across the 400 ice fields, the ice floes had a mean effective width of 8.39 m (SD = 4.68 m) and a mean area of 92.29 m$^2$ (SD = 231.34 m$^2$). The number of floes per ice field ranged from 300 at low concentration (20\%) to 1300 at high concentration (50\%).

\subsection{Vessel Model and Control System}
\label{app:vessel_model}
We used the full-scale PSV model from \cite{fossen1996identification, fossen2004} which has dimensions of 76.2~m $\times$ 18~m (L$\times$W) and weighs 6000 metric tons. For propulsion, the vessel is equipped with two main propellers and four tunnel thrusters (two aft and two fore), for a total of six actuators. The actuator configuration is illustrated in Fig.~\ref{fig:ship_model}. The dynamics of the vessel are described by a 3 DoF (i.e., surge $u$, sway $v$, and yaw rate $r$) linear state-space model intended for low-speed maneuvering on a horizontal plane: 
\begin{equation}
    \dot{\bm{\nu}} = \mathbf{A}\bm{\nu} + \mathbf{B}(\bm{\tau} + \bm{\tau}_{\text{env}})
\label{eq:dynamics}
\end{equation}
The model matrices $\mathbf{A}$ and $\mathbf{B} \in \mathbb{R}^{3\times 3}$ are taken from \cite{fossen1996identification, fossen2004}. In addition, we adapted the Dynamic Positioning (DP) based controller from \cite{fossen2004} to track a moving setpoint that traces the reference path at a target speed computed by the generated speed profile. This profile was defined as a linear ramp from the ship's current speed $U$ to the nominal speed $U_{\text{nom}} = 2$ m/s with constant acceleration 0.04 m/s$^2$. The DP controller uses pole placement for non-linear PD control, which computes the force vector $\bm{\tau}$ based on the error signals in surge, sway, and yaw. The propeller RPM commands for the six actuators are then calculated from $\bm{\tau}$ using 
the simple thrust allocation algorithm from \cite{fossen2004}. It uses the generalized inverse to solve for the control input $\bm{u} \in \mathbb{R}^6$ in the equation
\begin{equation}
    \bm{\tau} = \mathbf{T}\mathbf{K}\bm{u}.
\end{equation}
The diagonal matrix $\mathbf{K}\in\mathbb{R}^{6\times 6}$ is referred to as the \emph{thrust coefficient matrix} \cite{fossen2011handbook}. The main propellers are capable of producing a maximum of 799 kN of thrust each, while the tunnel thrusters each have a thrust limit of 200 kN. In the final step, we get the RPM commands $\{n_i\}_{i=1}^6$ where $n_i = \sign(u_i)\sqrt{|u_i|}$.

Note that the vessel model from \cite{fossen1996identification, fossen2004} lacks details on the shape and geometry of the vessel. As a result, we used the bow geometry for an ice-going supply vessel from \cite{daley2014gpu} to define our ship footprint as shown in Figure \ref{fig:ship_model}. 

\begin{table}[!t]
\centering
\begin{tabular}{ @{} l  l c @{}  }
\toprule
 Parameter & Value & Reference \\
\midrule
Physics engine time step $\Delta t_{\text{sim}}$ (s) & 0.005 & -\\
Controller sampling time $\Delta t_{\text{ctrl}}$ (s) & 0.02 & \cite{fossen2004}\\
Channel length (m) & 1000 & - \\
Channel width (m) & 200 & \cite{daley2014gpu} \\
Ship mass (kg) & $\text{6000}\cdot\text{10}^3$ & \cite{fossen2004} \\
Ship dimensions (L$\times$W) (m) & 76.2$\times$18 & \cite{fossen1996identification} \\
Ship target speed (m/s) & 2 & \cite{fossen2004} \\
Ship-ice friction coefficient & 0.05 & \cite{huang2020ship}\\
Ice-ice friction coefficient & 0.35 & \cite{huang2020ship} \\
Ship-ice restitution coefficient & 0.1 & \cite{yulmetov2016planar}\\
Ice-ice restitution coefficient & 0.1 & \cite{yulmetov2016planar}\\
Form drag coefficient $C_d$ & 1.0 & \cite{su2019numerical}\\
Angular velocity decay & 0.03 & -\\
Water density $\rho_w$ (kg/m$^3$) & 1025 & \cite{lubbad2018overview}\\
Ice density $\rho_{ice}$ (kg/m$^3$) & 900 & \cite{lubbad2018overview} \\
Ice thickness $h$ (m) & 1.2 & \cite{daley2014gpu}\\
Ice size range (m$^2$) & 16--10,000 & \cite{steer2008observed} \\
Ice side count range & 5--20 & \cite{daley2012gpu} \\
\bottomrule
\end{tabular}
\caption{Physics and other parameters used for our simulator.}
\label{table:sim_parameters}
\end{table}

\section{Parameter Settings in AUTO-IceNav}
\label{app:param_settings_experiments}
\subsection{Simulation Experiments}
\label{app:param_settings_sim_experiments}
The parameters for the tracking time $\Delta t$ and the receding horizon parameter $\Delta h$ in Algorithm \ref{alg:GeneralPlanner} were set at 30\,s and 500 \,m, respectively. At 2\,m/s, the ship travels a distance of 60 m, or nearly one ship length, between planning iterations. 

\begin{figure}[t]
    \centering
    \includegraphics[width=\columnwidth]{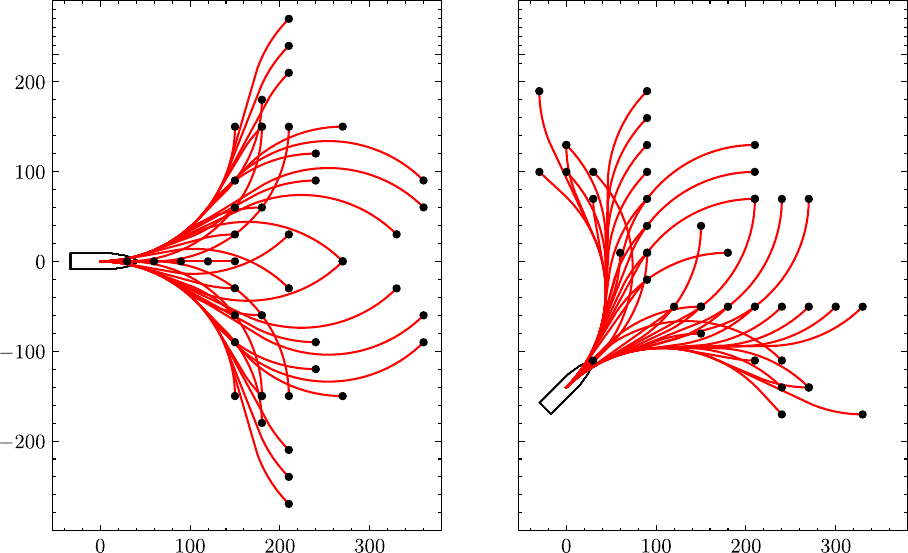}
    \caption{Primitives generated for the full-scale simulation experiments.}
    \label{fig:prims-sim}
\end{figure}

We set the resolution of the costmap to 2\,m $\times$ 2\,m. The configuration space $\mathcal{C}$ was discretized to 30~m $\times$ 30~m (roughly half a ship length) for the planar position, and the heading angle was discretized into 8 uniform intervals. The constraint on the yaw rate was established based on recommendations from a ship captain, who suggested 45 deg/min as an acceptable limit for low-speed maneuvering. At a nominal speed $U_{\text{nom}}$ of 2\,m/s, we get a minimum turning radius $r_{\min}$ of 150 m. The discretization of the configuration space and the minimum turning radius were used as inputs to the algorithm from \cite{botros2021multi} to generate our motion primitives. These included 45 motion primitives for even-numbered headings ($n \pi / 4$, where \( n \) is even) and 43 motion primitives for odd-numbered headings ($n \pi / 4$, where \( n \) is odd), as illustrated in Fig.~\ref{fig:prims-sim}.

To calibrate $\alpha$ in \eqref{eq:nav_framework_J_cost}, we compared our proposed navigation cost with the actual simulation cost in the Straight trials. Specifically, our goal was to match the ratio of total collision cost $C_f$ in \eqref{eq:swath_cost} to total cost $J$ with the ratio of kinetic energy loss $\Delta K_{\text{ship}}$ to total energy use $E$, meaning
\begin{equation}
    \frac{\alpha C_f}{L_f + \alpha C_f} =  \frac{\Delta K_{\text{ship}}}{E}.
\end{equation}
Solving for $\alpha$ gives us the expression,
\begin{equation}
    \alpha = \frac{\Delta K_{\text{ship}}}{E} \frac{L}{C_f\left(1 - \frac{\Delta K_{\text{ship}}}{E}\right)}.
\label{eq:alpha}
\end{equation}
We used \eqref{eq:alpha} to compute the weights $\{\alpha_i\}_{i=1}^{100}$, where each $\alpha_i$ corresponded to a sampled Straight trial. We then averaged these weights to obtain $\Bar{\alpha} = 4.8 \times 10^{-7}$, which we used as our calibrated $\alpha$ in the AUTO-IceNav trials.

Regarding the parameters for the ice concentration term in \eqref{eq:collision_cost2}, we set $\beta = 1$ and used a $51\times51$ kernel for the image convolution operation. For the optimization stage in Section \ref{sec:optim_step}, the set $\mathcal{B}_d$ consisted of 52 body points equally spaced along 4 rows (similar to Fig.~\ref{fig:body_points_options} (middle) with $\Delta b = 6$ m) where each body point was assigned the default body point weight \eqref{eq:default_body_point_weight}. The weight $\lambda$ for the smoothness term in \eqref{eq:optim_discrete} was set to $5 \times 10^4$ and the lattice planner solution was downsampled to an initial arc length step $\Delta s$ of 4 m. 

\subsection{Physical Experiments}
The parameters set in our framework were similar to the values used for the simulation experiments, and were scaled as necessary for the 20 m $\times$ 6 m ice channel. We increased the costmap resolution to 1/16 m $\times$ 1/16 m. At a nominal speed $U_{\text{nom}}$ of 0.2 m/s, we set a corresponding minimum turning radius of $r_{\min} = 1.5$ m. To generate the state lattice, we set the discretization for planar position to 0.75~m $\times$ 0.75~m, and we used the same discretization for heading as in the simulation setup. The algorithm from \cite{botros2021multi} was used here again to generate the motion primitives.

We deviated from the calibration method outlined in Appendix \ref{app:param_settings_sim_experiments} to set the parameter $\alpha$ in \eqref{eq:nav_framework_J_cost}. Preliminary OEB tests indicated that the ice provided limited resistance to ship motion. This meant that in terms of energy use, the optimal path was simply a straight path. As a result, we manually tuned $\alpha$ to apply a higher penalty on the total collision cost $C_f$, setting $\alpha = 10$. We set the horizon parameter $\Delta h$ to 6 m and the tracking duration parameter $\Delta t$ to 1 s in Algorithm \ref{alg:GeneralPlanner}.

\section{Trial Difficulty Evaluation}
\label{app:trial_difficulty}
A challenge encountered during the physical experiments was ensuring consistency between the three navigation approaches with respect to the initial arrangement of ice floes. Although periodic `resets' of the ice field were performed via manual control of the PSV, the starting ice field configuration remained a source of variation in trial performance. As a result, following the end of running experiments, each of the authors were tasked with assigning a subjective difficulty score for each of the trials. The evaluation process involved examining the first frame captured by the ceiling-mounted camera from a particular trial and responding to the following question:
\vspace{1em}
\\
\emph{On a scale from 1 to 5, where 1 is `Very Easy' and 5 is `Very Difficult,' how would you rate the overall difficulty of navigating the ship through the depicted ice field, considering the starting position of the ship and the characteristics of the ice field captured in the provided image?}
\vspace{1em}
\\
The evaluations were carried out independently, with the trial sequence randomized, and the navigation information hidden. 
Overall, the evaluations were generally consistent, with an average difference of 1.0 ± 0.7 between the highest and lowest assigned difficulty scores within a trial.
These scores yielded a mean difficulty for each trial and we binned the scores into three difficulty levels: \emph{easy}, \emph{medium}, and \emph{hard}, with intervals $[1, 2.5]$, $(2.5, 3.5)$, and $[3.5, 5]$, respectively. A summary of the trials per difficulty level is given in Table \ref{table:planner_difficulty_count}.

\begin{table}[t]
\centering
\begin{tabular}{ @{} l  c  c  c @{}  }
\toprule
\multicolumn{1}{c}{} & \multicolumn{3}{c}{Difficulty level} \\
\cmidrule{2-4}
 Method & Easy & Medium & Hard \\
\midrule
Straight & 8 & 5 & 7 \\
Skeleton & 3 & 8 & 9 \\
% Lattice  & 3 & 14 & 3 \\  % lattice2
AUTO-IceNav  & 6 & 7 & 7 \\     % lattice1
\bottomrule
\end{tabular}
\caption{Number of trials for each method and difficulty level.}
\label{table:planner_difficulty_count}
\end{table}

\section*{Acknowledgment}
The authors thank Jason Mills, Grant Hickey, Jason Murphy, and Derek Butler for their help during the experiments at the NRC OEB research facility. The authors also thank Thor I. Fossen for his support with the vessel dynamics and controller code from the MSS (Marine Systems Simulator) toolbox used in our simulator.

%%%%%%%%%%%%%%%%%%%%%%%%%%%%%%%%%%%%%%%%%%%%%%%%%%%%%%%%%%%%%%%%%%%%%%%%%%%%%%%%

\bibliographystyle{IEEEtran}
\bibliography{bib}

\end{document}